\documentclass[journal]{IEEEtran} 

\usepackage{graphics} % for pdf, bitmapped graphics files
\usepackage{epsfig} % for postscript graphics files
\usepackage{mathptmx} % assumes new font selection scheme installed
\usepackage{times} % assumes new font selection scheme installed
\usepackage{amsmath} % assumes amsmath package installed
\usepackage{amssymb}  % assumes amsmath package installed

\usepackage{wrapfig}
\usepackage{booktabs}
\usepackage{mathtools}
\usepackage{pgfplots}
\usepackage{tikz}
\usetikzlibrary{patterns}
\usepackage[noline,linesnumbered]{algorithm2e}

\usepackage{cite}
\usepackage{multicol}
\usepackage{placeins}
\usepackage[caption=false,font=footnotesize]{subfig}
\usepackage{adjustbox}
\usepackage{diagbox}
\usepackage[hidelinks]{hyperref} % to have hyperlinks and for the \ref macro
\usepackage{todonotes}
\usepackage[nolist]{acronym} % acronyms by the \ac{label} command
\usepackage{textcomp}
\newacro{ai}[AI]{Artificial Intelligence}
\newacro{dl}[DL]{Deep Learning}
\newacro{dnn}[DNN]{Deep Neural Network}
\newacro{dof}[DoF]{Degrees-of-Freedom}
\newacro{mcmc}[MCMC]{Markov Chain Mote Carlo}
\newacro{rl}[RL]{Reinforcemnt Learning}
\newacro{gp}[GP]{Gaussian Process}
\newacro{dof}[dof]{degree of freedom}
\newacroplural{dof}[dof]{degrees of freedom}

\newcommand{\figref}[1]{\hyperref[#1]{Fig.~\ref*{#1}}}
\newcommand{\tabref}[1]{\hyperref[#1]{Table~\ref*{#1}}}
\newcommand{\secref}[1]{\hyperref[#1]{Section~\ref*{#1}}}
\newcommand{\algoref}[1]{\hyperref[#1]{Algorithm~\ref*{#1}}}

\newlength\myindent
\def\panda{\textit{Franka Emika Panda }}

\def\ie{\textit{i.e.,}}
\def\eg{\textit{e.g.,}}

\def\pc{point-cloud}

\definecolor{significant}{RGB}{217,217,217}
\definecolor{verysignificant}{RGB}{255,255,255}
\begin{document}

\title{POMDP Manipulation Planning under Object Composition Uncertainty}

\author{Joni Pajarinen$^{1,2}$, Jens Lundell$^1$ and Ville Kyrki$^1$% <-this % stops a space
  \thanks{This work was supported by the Academy of Finland, decision 271394 and German Research Foundation (DFG) project PA 3179/1-1 (ROBOLEAP).}% <-this % stops a space
  \thanks{$^1$J.~Pajarinen, J.~Lundell and V.~Kyrki are with the Department of Electrical Engineering and Automation, Aalto University, Finland. $^2$J.~Pajarinen is with Intelligent Autonomous Systems, TU Darmstadt, Germany.}%
  \thanks{E-mail: {\tt\footnotesize Joni.Pajarinen@aalto.fi}, {\tt\footnotesize Jens.Lundell@aalto.fi},\break{\tt\footnotesize Ville.Kyrki@aalto.fi}}%
}

\markboth{}{Pajarinen \MakeLowercase{\textit{et al.}}: POMDP Manipulation Planning under Object Composition Uncertainty}

\maketitle

%%%%%%%%%%%%%%%%%%%%%%%%%%%%%%%%%%%%%%%%%%%%%%%%%%%%%%%%%%%%%%%%%%%%%%%%%%%%%%%%
\begin{abstract}
  Manipulating unknown objects in a cluttered environment is difficult
  because segmentation of the scene into objects, that is, object
  composition is uncertain. Due to this uncertainty, earlier work has
  concentrated on either identifying the ``best'' object composition
  and deciding on manipulation actions accordingly, or, tried to
  greedily gather information about the ``best'' object
  composition. Contrary to earlier work, we 1) utilize different
  possible object compositions in planning, 2) take advantage of
  object composition information provided by robot actions, 3) take
  into account the effect of different competing object hypotheses on
  the actual task to be performed. We cast the manipulation planning
  problem as a partially observable Markov decision process (POMDP)
  which plans over possible hypotheses of object compositions. The
  POMDP model chooses the action that maximizes the long-term expected
  task specific utility, and while doing so, considers the value of
  informative actions and the effect of different object hypotheses on
  the completion of the task. In simulations and in experiments with
  an RGB-D sensor, a Kinova Jaco and a Franka Emika Panda robot arm, a
  probabilistic approach outperforms an approach that only considers
  the most likely object composition and long term planning
  outperforms greedy decision making.
\end{abstract}


\IEEEpeerreviewmaketitle

\section{INTRODUCTION}

Service robots in domestic environments need the ability to manipulate
objects without good prior models in order to cope with the
variability of such environments. This need is usually approached by
attempting to model the objects on-line using sensors based on
stereopsis or structured light. When multiple measurements can be
acquired around an isolated object, this approach works quite
satisfactorily as the generated 3-D models can often be used for
successful manipulation. 

In cluttered scenes with multiple unknown objects, the segmentation of
objects, also known as object discovery in perception research,
becomes a major problem. Typically, the problem is to decide which of
the segments in an oversegmented scene belong to the same object. This
is challenging especially because objects can be partially occluded by
others. A promising approach towards solving object discovery is
interactive perception~\cite{bohg2017interactive}, where the object
configuration is examined actively, typically by manipulating the
objects and observing the results. Another line of work is to use
learned priors to find the most likely object composition. Despite the
recent advances, manipulation of unknown objects in cluttered
environments is still an open problem.

This paper proposes a solution to manipulation planning, which plans
over hypotheses of possible object compositions (segmentations)
instead of trying to determine a single best hypothesis. For
estimating the distribution of object compositions we propose a Markov
chain Monte Carlo (MCMC) procedure that produces approximately exact,
independent samples from the distribution. The approach combines
earlier ideas of interactive perception and learned composition priors
in a planning under uncertainty framework. The manipulation planning
problem is cast as a partially observable Markov decision process
(POMDP), which integrates active exploration and planning. This allows
us to take advantage of information provided by robot actions, plan
actions under object composition uncertainty into the future, and thus
tackle manipulation in challenging environments. In contrast to
earlier work, our approach 1) utilizes different possible object
compositions in decision making, 2) takes into account the effect of
competing hypotheses on the goal task, and 3) actively explores the
hypothesis space if that benefits the task. This journal article
extends our earlier conference papers \cite{pajarinen2015decision} and
\cite{pajarinen2014robotic} (i) by combining the MCMC procedure for
belief estimation in \cite{pajarinen2015decision} with the POMDP
planning in \cite{pajarinen2014robotic}; (ii) by including fully
hidden objects into the POMDP model by taking advantage of hidden
volume information; (iii) by a high volume of simulations for
statistically significant evaluations; (iv) new results using a Franka
Panda robotic arm verifying the general applicability of the approach.

This paper begins by surveying work related
to object discovery, interactive perception and planning under
uncertainty in Section \ref{sec:related_work}. Our framework of
manipulation planning over object compositions is then introduced in
Section \ref{sec:manipulating_object_compositions}. Section
\ref{sec:belief_estimation} proposes our approach for estimating the
distribution over object hypotheses. The state space of hypotheses is
then used for manipulation planning as described in Section
\ref{sec:manipulation_planning}. Experiments with two different
physical robot arms and an RGB-D sensor presented in Section
\ref{sec:experiments} demonstrate that the proposed approach is able
to integrate perception to the manipulation task and that the use of
multiple hypotheses improves system performance when compared to
considering only the most likely hypotheses. Section
\ref{sec:conclusions} concludes the paper.

\section{RELATED WORK}
\label{sec:related_work}

Our work focuses on manipulation under object composition
uncertainty. Below, we discuss related work w.r.t.\ image segmentation
and object discovery since we construct object hypotheses from RGB-D
data, active and interactive perception since our approach also
gathers information actively when needed, grasping unknown objects,
and manipulation under uncertainty.

\textbf{Image segmentation.} 
Image segmentation is a widely and
actively studied research field
\cite{haralick85,pal93,shi00,felzenszwalb04,richtsfeld14}. Most of the
earlier work concentrates on segmenting grey and color 2D images
\cite{haralick85,pal93}, but with the introduction of new cheap 3D-sensors such as Microsoft's Kinect or Intel's RealSense, research on
segmenting 3D images has gained in popularity
\cite{lai11,mishra12,richtsfeld12,richtsfeld14,danielczuk2019segmenting,xie2020best}. The 3D information from these sensors are especially useful for robotics tasks including grasping and manipulation \cite{morrison2018closing,mousavian20196,mahler2017dex,varley2017shape,xie2020best}. However, all of these methods consider only point estimates of the segmentation which, in case of poor segments, can lead to unsuccessful grasps \cite{xie2020best}.

To mitigate the influence of wrong segmentations, related works have thus focused on utilizing a probability distribution over segmentations \cite{beale11,vanhoof14} or considering many segmentation hypotheses \cite{katz13}. By treating the problem as such it is possible to use robots to gather information about the most likely segmentation through exploratory actions such as poking objects \cite{vanhoof14,katz13}. In this work we also treat the problem probabilistically but instead of considering only distributions over segmentations we model the complete probability distribution over object compositions and use this information for downstream decision making tasks.

A common technique for finding the best object composition is through
graph cuts \cite{shi00,boykov01,richtsfeld14}. In this paper, we
estimate the probability distribution over object compositions using a
Markov chain Monte Carlo (MCMC) procedure. Prior work on applying MCMC
to find a single best segmentation exists; for example, in the task
of segmenting humans from video frames using human shape models
\cite{zhao03}, or for segmenting 2D images \cite{tu02}. Due to
the inherent ambiguity in segmentation the authors also present in
\cite{tu02} a technique for selecting a fixed amount of distinct 2D
segmentations, instead of only the most likely segmentation.

\textbf{Object discovery.} Scene segmentation is a classic problem in
computer vision tightly coupled to object
recognition~\cite{yakimovsky1976boundary,redmon2016you} so that it can
be argued that the segmentation problem does not have unique solutions
if the objects are not known. Nevertheless, there is a need to
discover objects from scenes even when the objects are not known in
advance. Recent works in the area are based on learning to detect
objects based on synthetic training
data~\cite{danielczuk2019segmenting}, or often based on learning
general models used to recognize object classes from segments
(e.g. segment labeling in \cite{koppula11}), to detect segments based
on their ``objectness'' \cite{Karpathy-icra2013}, or to choose which
segments belong to a single object
\cite{richtsfeld12,richtsfeld14}. Our work follows the line of work of
\cite{richtsfeld12,richtsfeld14} but instead of trying to find a
single optimal composition as in
\cite{richtsfeld12,richtsfeld14,danielczuk2019segmenting}, considers
the distribution of possible compositions.

\textbf{Active and interactive perception.} Instead of a passive
approach object discovery can be approached from the point of view of
active perception \cite{bajcsy88}. Gaze control and foveation, which
are purely perceptual processes, have been proposed for object
discovery \cite{bjorkman10}. Furthermore, interactive
perception~\cite{bohg2017interactive} has been proposed as a promising
solution for object discovery with the goal of singulating
\cite{chang12}, clearing \cite{katz13}, or segmenting
\cite{van2014probabilistic} a pile of objects. These approaches use
poking or pushing actions to estimate the object composition. This
paper follows the interactive perception paradigm but in contrast to
the works above, integrates the perception with goal-directed planning
so that perceptual actions are only used when they are expected to
support the task goal.

\textbf{Grasping unknown objects.}
Grasping unknown objects has got significant attention in the research
community, especially after Saxena's work
\cite{Saxena-nips2006} which proposed the use of machine learning for
planning good grasps. Since then, many new data-driven methods have
arisen
\cite{Lenz-rss2013,mahler2017dex,satish2019policy,morrison2018closing,mousavian20196,qin2020s4g,mahler2019learning}. Out
of these the most influential works, \eg~DexNet
\cite{mahler2017dex,mahler2019learning} and 6-DOF GraspNet
\cite{mousavian20196}, are based on deep-learning and propose grasps
directly from raw sensor readings such as depth-images or
point-clouds. In order to reach a high grasp success rate grasps are
often constrained to top-down and to 4 degrees of freedom
\cite{mahler2017dex,satish2019policy,morrison2018closing}. In this
work we also consider top-down grasps but, instead of using a
data-driven method, we align the robot hand according to the principal
axes of the \pc. In addition, we consider grasping as a component for
both informative and goal-directed actions; hence, even failed grasps
give valuable information about object composition that is used for
planning future grasps.

\textbf{Manipulation planning under uncertainty.} In planning
manipulation actions under uncertainty, such as planning where to grasp an
object, classical deterministic planning can be used to reduce
uncertainty. For example, Dogar et al.\ \cite{dogar12} plan pre-grasp
pushing actions that collapse pose uncertainty of a target before
executing a grasping action. This type of approach is usually
only available for completely known objects. 

When faced with limited knowledge, POMDP-like approaches can be used
to plan over a distribution of states. Hsiao et
al.\ \cite{Hsiao-icra2007} proposed the partitioning of a
one-dimensional configuration space to yield a discrete POMDP which
can be solved for an optimal policy. In planning grasp locations, the
state-of-the-art includes probabilistic approaches with a short time
horizon. The goal can be formulated either as positioning the robot
accurately as in \cite{Hsiao-ar2011} or maximizing the probability of
a successful grasp as in \cite{Laaksonen-iros2012}. The short-term
planning can also be extended to include information gathering actions
\cite{Nikandrova-ras2013}.

In POMDP based approaches for multi-object manipulation, Monso et
al.\ \cite{monso12} use a POMDP definition designed specifically for
clothes separation. In contrast to \cite{monso12}, this paper does not
assume that each object is uniform in color, but instead, complex
multi-colored and textured objects are considered for object
discovery. Moreover, the state space model of \cite{monso12} is
clothes separation specific, modeling the number of clothes in
different areas. Our approach reasons about objects directly.

Li et al.\ \cite{li2016act} formulate emptying a refrigerator as a
POMDP. Li et al.\ \cite{li2016act} do not explicitly model a probability
distribution over object hypotheses but assume a priori six different
types of known objects and perfect segmentation. Xiao et al.\ \cite{xiao2019online}
tackle object search using a POMDP model that also considers fully
occluded objects. Xiao et al.\ \cite{xiao2019online} assume perfect object
composition segmentation with some uncertainty in object
locations. Moreover, Xiao et al.\ \cite{xiao2019online} assume that the number and
models of objects are known in advance and use this information for
finding hidden objects. In contrast to
\cite{li2016act,xiao2019online}, our approach does not assume any
information on object models or number of objects. Instead, we plan
manipulation actions based on a probability distribution of object
compositions.

\section{Manipulating object compositions}
\label{sec:manipulating_object_compositions}
We consider the scenario of a robot manipulating unknown objects based
on RGB-D data. The manipulation goal is defined in terms of simple
features that can be observed incompletely from the point clouds. For
example, the goal could be to move all objects with a certain color to
a particular location. Manipulating unknown objects is difficult
because even if RGB-D data is available, the robot does not know in
advance the shape or color of the objects. Thus, the robot has to
guess which parts of the point cloud belong to the same
object. Occlusion and noisy sensor readings make this task
hard. Attempting to segment individual objects from the point cloud
typically results in oversegmentation, which leads to the problem of
deciding which segment belongs to which object, in other words,
forming \emph{object hypotheses}.

In previous works such as \cite{Fischinger-icra2013}, the choice of an
action is based on the most probable hypothesis of object
composition. The shortcoming of this approach is that it does not take
into account the long term effects of uncertainty or the value of
information gathering actions. Instead, we propose to choose the action
that maximizes long term reward over the current and future
distribution of possible object compositions. By considering a
temporally evolving system, the robot can infer from past grasp
attempts the likelihood of object hypotheses.

\begin{figure*}[tb]
    \centering
    \includegraphics[width=\textwidth]{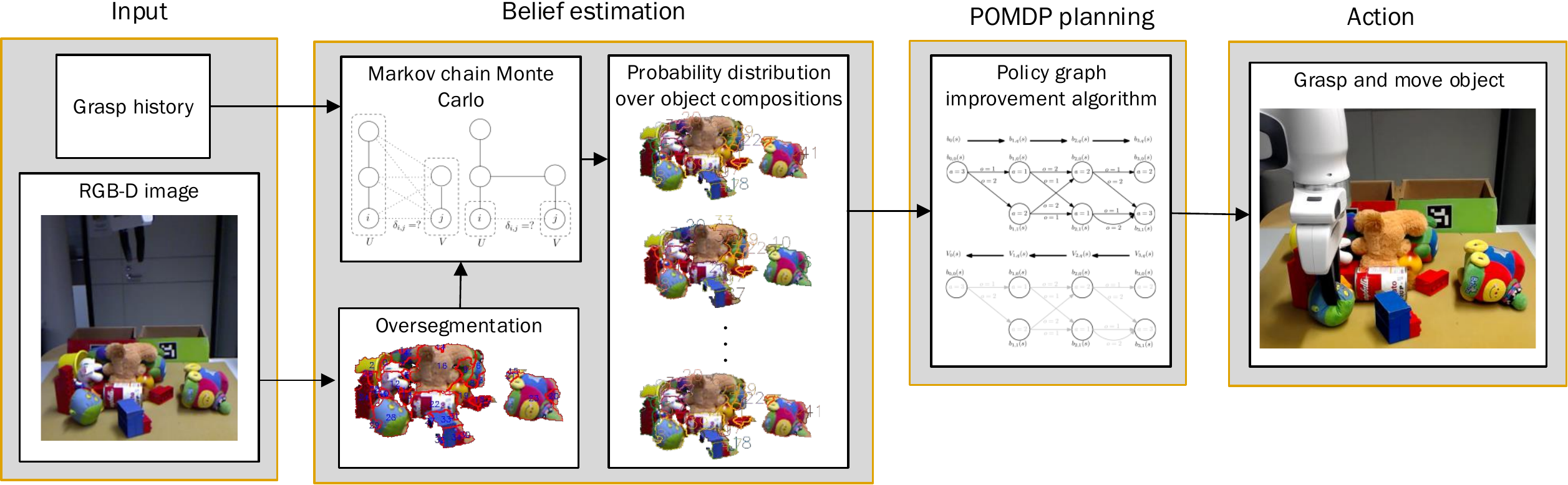}
    \caption{Overview of the proposed approach. At each time instant
      the robot obtains and oversegments RGB-D scene data. From
      pair-wise segment probabilities of belonging to the same object
      we generate a probability distribution over possible object
      compositions. We condition the probability distribution on
      past grasp successes and failures (Section
      \ref{sec:belief_estimation}). The robot uses a POMDP to select
      the best long-term manipulation action for the current object
      distribution (Section \ref{sec:manipulation_planning}) and
      executes the action. For segmenting RGB-D data and estimating
      probabilities for segment pairs, we use \cite{richtsfeld14}.}
    \label{fig:approach_overview}
\end{figure*}

Fig.~\ref{fig:approach_overview} shows an overview of the proposed
approach. At each time step we capture an RGB-D image by a vision
sensor such as a Microsoft Kinect and oversegment the RGB-D
image. Further, using Markov chain Monte Carlo we generate a
probability distribution over possible object compositions where each
composition consists of segment patches while taking into account
previous manipulation outcomes. We perform long-term planning over
possible future object composition distributions using a POMDP
model. Finally, the best POMDP action is executed on the robot.
Before going into belief estimation and manipulation planning details,
we describe first how to model robotic manipulation as a POMDP.

\subsection{Robotic manipulation as a partially observable Markov decision process (POMDP)}

A POMDP defines optimal behavior for an agent in an uncertain world
with noisy, partial measurements, when the stochastic world model is
accurate and when the agent's goal has been defined
precisely. Previously, POMDPs have yielded good results in robotic
applications such as navigation~\cite{koenig1998xavier}, autonomous
driving~\cite{bai2015intention}, human-robot
interaction~\cite{taha2011pomdp} and
manipulation~\cite{Hsiao-icra2007,li2016act,xiao2019online,monso12,pajarinen2017robotic}. We
utilize a POMDP because it takes uncertainty in action effects and
observations into account. Moreover, a POMDP assigns the correct
long-term value to informative actions which are needed when exploring
object hypotheses.

The temporal model of a POMDP is defined by the transition probability
$P(s^{\prime} | s, a)$ from state $s$ to the next time step state
$s^{\prime}$, when action $a$ is executed, and the probability $P(o |
s^{\prime}, a)$ of observing $o$, when action $a$ was executed and the
world moved to the state $s^{\prime}$. A real-valued reward $R(s, a)$
for executing action $a$ in state $s$ encodes the objective. An
optimal policy $\pi$ maximizes the expected reward $E\left[
  \sum_{t=0}^{T-1} R(s(t), a(t)) | \pi, b_0 \right]$ over $T$ time
steps, where $b_0$ denotes the initial \emph{belief}, a
probability distribution over world states. At each time step, the
agent decides on an action $a$ based on its current belief. In
principle, the belief can be kept up-to-date given an accurate
temporal model. However, because an accurate model is in practice not
available, we instead estimate the belief at each time step from
current visual sensor data and past history, and use the \emph{online}
POMDP method called particle based policy graph improvement algorithm
(PPGI) introduced in \cite{pajarinen2017robotic} (the technical
report~\cite{pajarinen2020technical} provides further details) to
compute a new policy. To cope with a huge state space, PPGI uses a
state particle representation for the belief $b(s)$.

Our POMDP model uses a probability distribution over possible object
hypotheses. Grasping actions occur in the space of object
hypotheses. Shortly, the probability of successfully grasping a
hypothetical object, and observing its attributes (for example color),
depend on how occluded the object is
\cite{pajarinen2017robotic}. Moreover, we assume that a previously
failed grasp can not succeed unless the occlusion on the grasped
object changes. Formally, the POMDP state $s = (s_1, s_2, \dots, s_N)$
is a combination of object states $s_i = (s_i^{\textrm{loc}},
s_i^{\textrm{attr}}, s_i^{\textrm{hist}})$ where $s_i^{\textrm{loc}}$
is the semantic location such as ``on top of the table'' or ``in a
box'', $s_i^{\textrm{attr}}$ attributes such as object color, and
$s_i^{\textrm{hist}}$ historical information for object $i$. The
number of objects $N$ and the objects $s_i$ themselves differ between
different POMDP states since we model a variety of object hypotheses
as discussed in more detail next.

\section{Belief estimation}
\label{sec:belief_estimation}

The belief consists of state particles and their probabilities. Here,
instead of objects, each state consists of a set of \emph{object
  hypotheses} called an \emph{object composition}. An object
hypothesis consists of a set of segments, where every segment is
connected either directly or in-directly to each other. The segments
are not connected to segments outside this set. Two segments can be
directly connected if they occlude each other or if another object
occludes both (the direct connection is then behind this occluding
object).

The probability of a sampled state is proportional to the probability
of the related object composition to exist. The probability for an
object composition to exist depends on the history of successful and
failed grasps. The key insight is that previously performed grasps
must have failed for an object hypothesis, otherwise the object would
have been moved. Furthermore, a grasp can only succeed for an
incorrect hypothesis, when the incorrect hypothesis is part of a
hypothesis, for which the grasp succeeds.

\begin{figure}[th]
  \centering
  \includegraphics[width=0.4\textwidth]{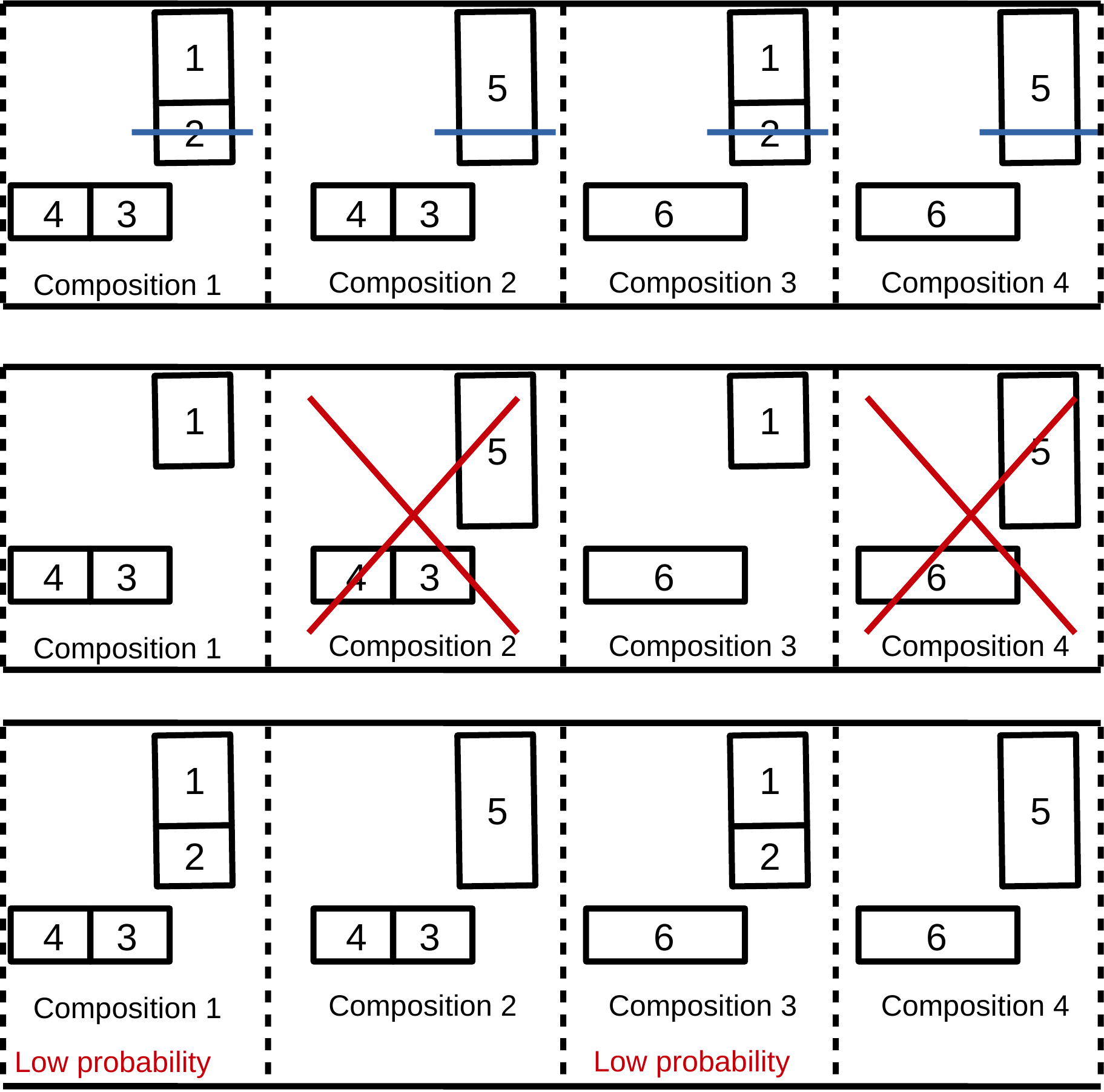}
  \caption{Illustration of conditioning on grasp outcome. Here the
    probability distribution over object compositions consists of four
    possible compositions with six possible objects. \textbf{(Top)}
    The robot executes blue grasp that grasps object
    2. \textbf{(Middle)} The grasp succeeds and object 2 is
    removed. Compositions 2 and 4 which did not contain object 2
    conflict with reality and are eliminated. \textbf{(Bottom)} The
    grasp fails either due to object 2 not existing or a bad
    grasp. Therefore, probability of object compositions that contain
    object 2 decreases.}
  \label{fig:condition_on_grasp_success}
\end{figure}

To estimate the belief, we segment the observed point cloud and
compute the connectedness probability for each segment pair. Based on
these probabilities, and whether segments can be directly
connected, we define a Markov chain which converges to a distribution
over object compositions. We sample object compositions from this
Markov chain after a burn-in period. A belief state corresponds to a
sampled object composition with sampled object attributes. The
probability of the belief state is set proportional to the probability
of the sampled object composition, which is computed (for uniform
priors) as the probability of the observation/action history
conditional on the object composition. This means that the belief over
object compositions is shaped by past events: for example, if the
robot fails to grasp an object hypothesis, which should be easy to
grasp when the object hypothesis is correct, then the hypothesis is
likely incorrect, and the belief will reflect
this. Fig.~\ref{fig:condition_on_grasp_success} illustrates how grasp
successes and failures influence the probability distribution over
object compositions. Next, we will discuss how to sample object
compositions and then show how to estimate the conditional probability
of an object composition given past events.

\subsection{Markov chain sampling of object compositions}

To generate object compositions in a computationally efficient way, we
first segment the image into pixel patches or segments as shown in
Fig.~\ref{fig:approach_overview}, and then combine the segments into
object compositions which consist of object hypotheses. Previous work
on segmenting an image and then combining the segments into a single
object composition immediately \cite{richtsfeld14}, or through
interaction \cite{beale11,vanhoof14}, exists. We instead maintain a
probability distribution over object compositions and make decisions
based on the probability distribution.

More formally, denote with $\delta_{i,j}$ whether segments $i$ and $j$
are directly connected: $\delta_{i,j} = 1$ and $\delta_{i,j} = 0$
denote direct and no direct connection, respectively. Direct
connection means segments are assumed physically connected. Denote
with $\mathbf{\delta}$ all possible direct connections. Denote with
$\mathbf{h} = (h_1,\dots,h_N) \in \mathcal{H}$ an object composition,
where $h_k$ is an object hypothesis and $\mathcal{H}$ the space of
object compositions. An object hypothesis $h_k$ is a set of segment
indices where all index pairs $i \in h_k$, $j \in h_k$ are connected
either directly or indirectly through other segments, denoted with
$c_{i,j}(h_k) = 1$ for all index pairs $i \in h_k$, $j \in h_k$ and
$c_{i,j}(h_k) = 0$ otherwise. Note that our sampling procedure in
Algorithm~\ref{alg:sample_object_composition} in
Section~\ref{sec:MCMC} uses direct and indirect connections to
estimate the probability of a segment pair connection.

In the worst case, the dimensionality of $|\mathcal{H}|$ is $2^{N^2/2}$
w.r.t.\ the number of segments $N$. In practice, $|\mathcal{H}|$ is
usually lower since segments with only ``air'' between them cannot be
directly connected. Often, in real-world scenes, $|\mathcal{H}|$ is a
product of the dimensionality of disconnected groups of segments
$\prod_i 2^{N_i^2/2}$, where $N_i$ is the number of segments in a
segment group. For exact computation this is still intractable, and
therefore we use an approximate particle representation for the
probability distribution over object compositions: $P(\mathbf{h}) =
\sum_i w_i \mathbf{h}_i$, where $\sum_i w_i = 1$ and $w_i \geq 0 \;
\forall i$.

\subsection{Markov chain Monte Carlo}
\label{sec:MCMC}

In order to generate the particle based probability distribution over
object compositions which can be used as a basis for decision making, we utilise Gibbs sampling (also known as Glauber
dynamics) \cite{casella92,mackay03,levin09}. We randomly sample direct
connections one connection at a time. We will first discuss how a new
Markov chain state is sampled, then show that the proposed Markov
chain is ergodic and converges to a unique distribution for
non-deterministic connection probabilities, and finally present a
sampling procedure that aims to generate exact, independent samples.

\begin{figure}
	\centering
	\vspace{0.4em}
	\begin{tabular}{cc}
		\fbox{\includegraphics[width=3.4cm]{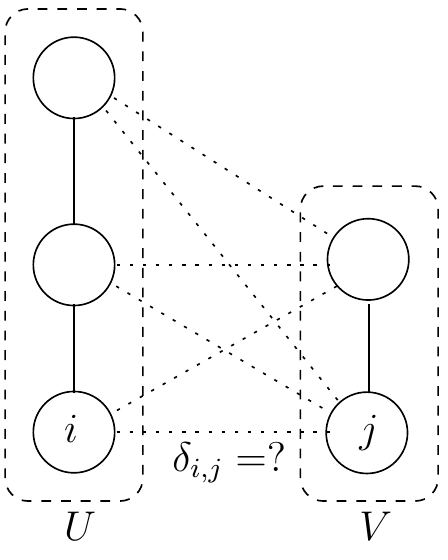}}&
		\fbox{\includegraphics[width=3.4cm]{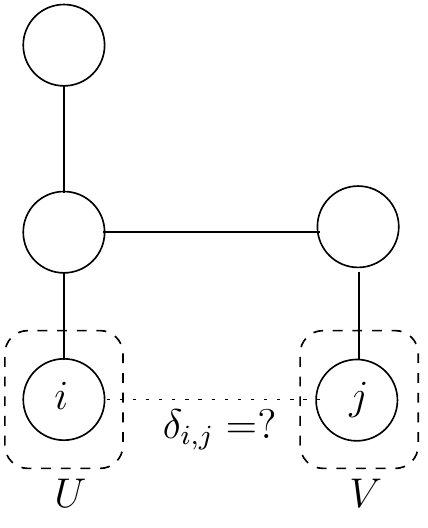}}\\
		(a) & (b) 
	\end{tabular}
	\caption{Effect of indirect connections on the connection
		probability between segments $i$ and $j$. A circle denotes a
		segment and a solid line denotes a connection between
		segments. Dotted lines denote which connection probabilities are
		used for sampling the connection between $i$ and $j$. $U$ and $V$
		denote the sets of segments directly or indirectly connected to
		$i$ and $j$, respectively. (a) Because $i$ and $j$ are not
		indirectly connected we have to consider the connection
		probabilities between segments that will become part of the same
		object, that is, we have to take into account the connection
		probabilities between all segments in the sets $U$ and $V$. (b)
		Because $i$ and $j$ are already indirectly connected we consider
		only the probability of the direct connection between $i$ and
		$j$.}
	\label{fig:connection_probabilities}
\end{figure}

Our sampling technique for generating a new Markov chain state takes
advantage of the fact that evaluating the probability for a single segment
connection is fast because we only need to consider local segment
connections. The sampling technique consists of two steps: 1)
select randomly two segments $i$ and $j$ which may be directly
connected, 2) sample the direct connection from the probability
distribution, which is estimated by assuming the direct connection is
disabled and by keeping other direct connections fixed to their
current values. When $i$ and $j$ are indirectly connected, that is,
part of the same object through some other connections, the
probability for the direct connection between $i$ and $j$ depends only
on the prior probability of $i$ and $j$ being part of the same object
because connecting $i$ and $j$ would not change which object
hypothesis other segments would belong to. When $i$ and $j$ are not
already part of the same object, the probability for the direct
connection depends on the probabilities between the segment sets $U$
and $V$ which connecting $i$ and $j$ would connect into the same
object hypothesis. Fig.~\ref{fig:connection_probabilities} illustrates
this.

Algorithm~\ref{alg:sample_object_composition} defines formally how to
sample a new object composition $\mathbf{h^*}$, when given the current
object composition $\mathbf{h}$, uniformly randomly selected possible
direct connection index $w$, and finally the sampled value $q \sim
\textrm{Uniform}(0, 1)$. The algorithm first chooses the direction connection
candidate using $w$, then on lines \ref{alg:1:U} and \ref{alg:1:V}
determines the segment sets $U$ and $V$ which the direct connection
would connect. On lines \ref{alg:1:h1}, \ref{alg:1:h0}, the algorithm
computes the probability for the segment sets $U$ and $V$ to belong to
the same object hypothesis when $i$ and $j$ are connected and when
not. Assuming a uniform direct connection prior, line
\ref{alg:1:deltah} computes the direct connection probability, and
line \ref{alg:1:deltah} finally sets the direct connection on or off
using $q$.

\DontPrintSemicolon
\SetKwProg{Fn}{}{}{}
\begin{algorithm}[tb]
  \SetKwFunction{Sample}{$\mathbf{h^*}$ = Sample}
  \Fn{\Sample{$\mathbf{h}$,$w$,$q$}}
  {
    \KwIn{Composition $\mathbf{h}$, random values $w$ and $q$}
    \KwOut{New composition $\mathbf{h^*}$}
    $\delta_{i,j} \gets$ The $w$th direct connection\;
    $\mathbf{h^*} \gets \mathbf{h}$ so that $\delta_{i,j} = 0$\;
    $U \gets
    \begin{dcases}
      i & \text{if } c_{i,j}(h^*) = 1 \\
      i \cup \left\{{ u | c_{i,u}(h^*) = 1 } \right\} &
      \text{if } c_{i,j}(h^*) = 0
    \end{dcases}$\label{alg:1:U}\;
    $V \gets
    \begin{dcases}
      j & \text{if } c_{i,j}(h^*) = 1 \\
      j \cup \left\{{v | c_{j,v}(h^*) = 1}\right\} &
      \text{if } c_{i,j}(h^*) = 0
    \end{dcases}$\label{alg:1:V}\;
    $P(h^* | \delta^*_{i,j} = 1) \gets \frac{1}{Z(U,V,h^*)}
    \prod_{u \in U} \prod_{v \in V} P(c_{u,v}(h^*) = 1)$\label{alg:1:h1}\;
    $P(h^* | \delta^*_{i,j} = 0) \gets \frac{1}{Z(U,V,h^*)}
    \prod_{u \in U} \prod_{v \in V} P(c_{u,v}(h^*) = 0)$\label{alg:1:h0}\;
    $P(\delta^*_{i,j} = 1 | h^*) \gets
    \frac{P(h^* | \delta^*_{i,j} = 1)}{
      P(h^* | \delta^*_{i,j} = 1) + 
      P(h^* | \delta^*_{i,j} = 0)}$\label{alg:1:delta}\;
    $\delta^*_{i,j} \gets
    \begin{dcases}
      0 & \text{if } P(\delta^*_{i,j} = 1 | h^*) \leq q \\
      1 & \text{if } P(\delta^*_{i,j} = 1 | h^*) > q
    \end{dcases}$\label{alg:1:deltah}\;
  }
  \vspace{0.3em}
  \caption{Sample new object composition.}
  \label{alg:sample_object_composition}
\end{algorithm}

\paragraph{Ergodicity of the Markov chain} When the connection
probability $P(c_{i,j})$ for any two segment patches is
non-deterministic $0 < P(c_{i,j}) < 1$, the Markov chain generated by
Algorithm~\ref{alg:sample_object_composition} is ergodic and converges to
a unique distribution. Because $P(c_{i,j})$ is non-deterministic the
probabilities on lines \ref{alg:1:h1}, \ref{alg:1:h0}, and
\ref{alg:1:delta} are non-deterministic, and since we randomly
select the direct connection to consider,
Algorithm~\ref{alg:sample_object_composition} enables or disables any
direct connection with non-zero probability. Therefore, the Markov
chain is ergodic and converges to a unique distribution in the
limit. Note that due to the inherent uncertainty in segmentation
the condition $0 < P(c_{i,j}) < 1$ usually applies, e.g.\ in the
experiments in Section~\ref{sec:experiments}.

\paragraph{MCMC procedure} We would like our MCMC approach to
produce independent samples from the correct distribution. Therefore,
our MCMC approach first aims to get an exact sample, that is, a sample
from the correct probability distribution, then continue sampling
until having enough independent samples ($\mathbf{H}$ in
Algorithm~\ref{alg:sample_compositions}). Algorithm~\ref{alg:sample_compositions}
shows the proposed MCMC approach. Since the Markov chain state is a
discrete combination of binary variables, each variable denoting
whether two segments are directly connected, we base our approach on
the coupling from the past (CFTP) \cite{propp96} technique which aims
at providing exact samples. The basic idea of CFTP is to run Markov
chains starting from each possible state with the same random numbers,
starting further back in time, until the chains collapse (see
\cite{propp96} for why). For monotone \cite{propp96} and anti-monotone
\cite{haggstrom98} Markov chains only two starting states are
needed. However, our chain is not monotone nor anti-monotone. Due to
the large number of states we start CFTP from a limited dispersed
set of states: the all connected, all disconnected, and from a fixed
number of randomly selected states. We can make the collapsed sample
more likely to be exact by increasing the number of starting states:
when the starting states cover the whole state space the collapsed
sample will be exact \cite{propp96}. Algorithm~\ref{alg:CFTP} shows
the CFTP procedure we use. In the experiments, we used $100$
($N_{\textrm{START}}$ in Algorithm~\ref{alg:sample_compositions} and
Algorithm~\ref{alg:CFTP}) starting states.

After CFTP, we start the actual sampling from the collapsed sample,
and double the sampling horizon until having enough independent
samples. We use the minimum of the effective sample size (ESS)
\cite{gelman13} over all possible direct connections
($N_{\textrm{ESS}}$ in Algorithm~\ref{alg:sample_compositions}) as a lower
bound estimate for the number of independent samples. Sampling stops
when the estimate for independent samples is large enough. We also use
a hard limit on the number of generated samples ($T_{\textrm{MAX}}$ in
Algorithm~\ref{alg:sample_compositions}).

\begin{algorithm}[tb]
	\SetKwFunction{CallCFTP}{CFTP}
	\SetKwFunction{Sample}{Sample}
	\SetKwFunction{SampleCompositions}{$\mathbf{H} =$Compositions}
	\Fn{\SampleCompositions{$N_{\textsc{ESS}}$,$N_{\textsc{start}}$,$\mathbf{|H|}$}}
	{
		\KwIn{ESS target $N_{\textsc{ESS}}$}
		\KwOut{Compositions $\mathbf{H}$}
		$\left\{{\mathbf{h_1}, T}\right\} \gets $ \CallCFTP{$N_{\textsc{start}}$}\;
		$t \gets 1$, $\mathbf{H} \gets \mathbf{h_1}$\;
		\While{$((\textsc{ESS}_{min}(\mathbf{H}) < N_{\textsc{ESS}}) \textsc{ and}$\\
			\mbox{}\hphantom{while (}$(T < T_{\textsc{max}}))$}
		{
			\While{$t < T$}
			{
				$\mathbf{h_{t+1}} \gets $ \Sample($\mathbf{h_t}$,$w$,$u$)\;
				$\mathbf{H} \gets \left\{ {\mathbf{H}, \mathbf{h_{t+1}}} \right\}$\;
				$t \gets t + 1$\;
			}
			$T \gets 2 T$\;
		}
		$\mathbf{H} \gets $ Prune $\mathbf{H}$ evenly to size $\mathbf{|H|}$
	}
	\vspace{0.3em}
	\caption{Sample a set of object compositions.}
	\label{alg:sample_compositions}
\end{algorithm}

\SetAlCapHSkip{0em} 
\begin{algorithm}[tb]
	\SetKwFunction{CFTP}{$\{\mathbf{H_T}, T\}$ = CFTP}
	\Fn{\CFTP{$N_{\textsc{start}}$}}
	{
		\KwIn{\# of start compositions $N_{\textsc{start}}$}
		\KwOut{Composition $\mathbf{H_T}$ at time $T$}
		$\mathbf{H_{\textsc{init}}} \gets 
		\{ \{\mathbf{h_1}| \mathbf{\delta}=1\},
		\{\mathbf{h_2}| \mathbf{\delta}=0\},$\\
		\mbox{}\hphantom{$\mathbf{H_{\textsc{init}}}$a}
		$\{\mathbf{h_3},\dots,\mathbf{h_{N_{\textsc{start}}}}|
		\mathbf{\delta}=\text{random}\} \}$\;
		$T \gets 1$\;
		\Repeat{$|\mathbf{H_T}| = 1$}
		{
			$T \gets 2 T$, $\mathbf{H_T} \gets \mathbf{H_{\textsc{init}}}$\;
			$w_T,\dots,w_{T/2} \gets $ Random\;
			$u_T,\dots,u_{T/2} \gets $ Random\;
			\For{$t \gets T$ \textbf{to} 1}
			{
				$\mathbf{H_{T-t+1}} \gets \emptyset$\;
				\ForEach{$\mathbf{h_{T-t}} \in \mathbf{H_{T-t}}$}
				{
					$\mathbf{H_{T-t+1}} \gets \mathbf{H_{T-t+1}} \cup$\\
					\Indp\Sample{$\mathbf{h_{T-t}}$,$w_t$,$u_t$}\;
				}
			}
		}
	}
	\vspace{0.3em}
	\caption{Coupling from the past (CFTP).}
	\label{alg:CFTP}
\end{algorithm}

\subsection{Probability of an object composition given past events}

In general, the probability of an object composition $h =
(h_1,\dots,h_N)$, where $h_i$ is a single object hypothesis, depends
on the sequence of past actions and observations $\theta_t =
(a(0),o(1),a(1),o(2),\dots,a(t-1),o(t))$, where $t$ denotes the
current time step. We assume uniform priors, independent object
hypotheses, and independent history events:
\begin{equation}
  P(h | \theta_t) = \prod_{i=1}^N P(h_i | \theta_t)
  = \prod_{i=1}^N \prod_{k=1}^t P(a(k-1), o(k) | h_i).
\end{equation}
For object hypothesis manipulation, we maintain a history of unique
executed grasps. In our model, there is no need to remember multiple
identical grasps; a previously failed grasp cannot succeed again,
unless the occlusion of the object, for which the grasp is optimized,
changes because then the grasp also changes. The composition
probability conditional on past independent grasps
$(\textrm{grasp}_1,\dots,\textrm{grasp}_M)$ is
\begin{equation}
  P(h | \theta_t) = \prod_{i=1}^N \prod_{k=1}^M 
  P(\textrm{grasp}_k| h_i).
\end{equation}

\subsection{Hallucinating hidden objects}
\label{sec:hallucination}

\begin{wrapfigure}{R}{4.0cm}
  \centering
  %\vspace{0.4em}
  \includegraphics[width=4.0cm]{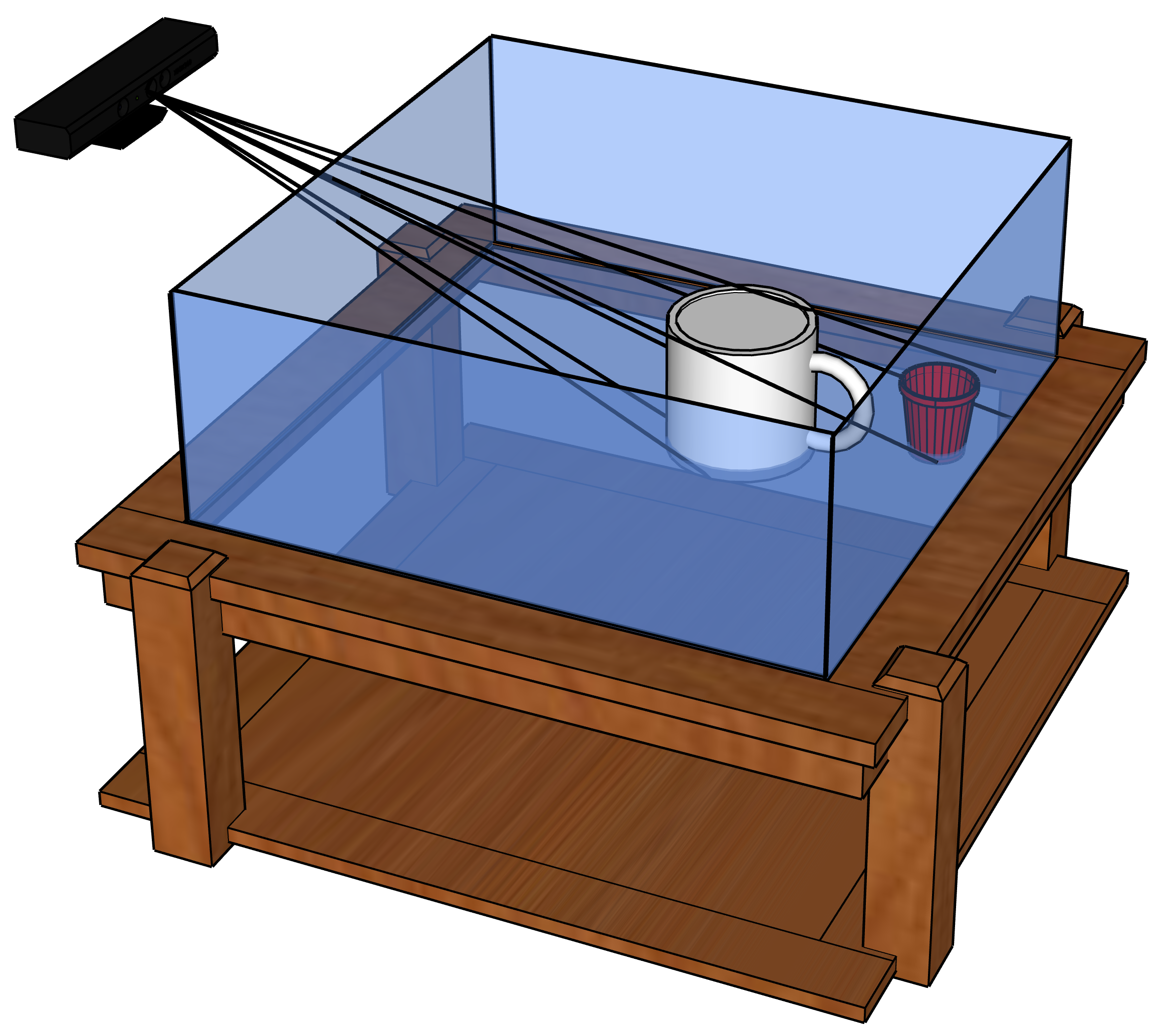}
  \caption{Illustration of hidden objects. The transparent ``glass''
    box denotes the workspace inside which the robot currently
    operates. A white cup fully occludes a red cup. To reason about
    hidden objects we use the insight that only a limited amount of
    objects can be hidden behind other objects. We assume that the
    probability for an object to be hidden behind another object
    depends on the volume hidden behind the occluding object, and, the
    object occupancy density.}
  \label{fig:hallucination}
\end{wrapfigure}
Due to occlusion the robot does not see the complete scene: objects
can hide behind other objects as shown in
Fig.~\ref{fig:hallucination}. In order to allow the robot to reason
about hidden objects we allow the robot to hallucinate hidden objects
into the object composition distribution. For hallucinating hidden
objects, we utilize the key insight that the probability of hidden
objects depends on the amount of hidden space, and, the object
occupancy density of that space. Using these assumptions we transform
the probability distribution over visible objects into a probability
distribution over both visible and hidden objects.

Let us denote with $V_{\textrm{visible}}$ visible volume and with
$V_{\textrm{hidden}}$ hidden volume, that is, how much workspace the
robot sees and does not see, respectively. To compute hidden volume
$V_{\textrm{hidden}}$, we project each visible voxel onto the
workspace boundary from the origin (camera location) and estimate the
volume between the rectangular visible voxel and the voxel projected
onto the workspace boundary. For each segment patch, we get the hidden
volume by summing over the segment voxels. Summing over all segment
patches yields then the hidden volume $V_{\textrm{hidden}}$. Let us
denote the average height of the visible segment patches with $d$, the
cube floor area with $A_C$, and the average number of objects with
$n_{\textrm{objects}}$. The minimum for $n_{\textrm{objects}}$ is set
to $1$. We estimate the total volume of interest $V_{\textrm{total}}$
for a cubical workspace as $V_{\textrm{total}} = 2 d A_C$. The visible
volume $V_{\textrm{visible}}$ is then total volume minus hidden
volume: $V_{\textrm{visible}} = V_{\textrm{total}} -
V_{\textrm{hidden}}$.

The object occupancy density is $p_{\textrm{objects}} =
n_{\textrm{objects}} / V_{\textrm{visible}}$. For each segment patch
$i$, we compute the expected number of hidden objects
$n^i_{\textrm{objects}}$ behind the patch, based on the object
occupancy density $p_{\textrm{objects}}$ and the patch's hidden space
volume $V^i_{\textrm{hidden}}$: $n^i_{\textrm{objects}} =
p_{\textrm{objects}} V^i_{\textrm{hidden}}$. In the experiments,
although the model allows for multiple hidden objects behind an
object, we assume for simplicity that there is at most one object
behind an occluding object: for each state particle, a hallucinated
object is sampled behind each object hypothesis according to the
probability $1 - \Phi\left((1 - n^i_{\textrm{objects}}) /
\sqrt{n^i_{\textrm{objects}} / 4}\right)$ which corresponds to the
Gaussian probability of having at least one hidden object when the
mean of the Gaussian is $n^i_{\textrm{objects}}$ and the standard
deviation $\sqrt{n^i_{\textrm{objects}} / 4}$. Depending on the task
we can estimate probabilities for hidden object properties based on
visible objects and a priori information. For example, in the object
search experiments we estimate the probability for hidden objects to
be red based on the average probability for redness. Now that we have
discussed how to estimate the belief over object compositions we will
next discuss how to plan manipulation actions using the belief.

\section{Manipulation planning }
\label{sec:manipulation_planning}

After the robot has estimated the current belief it decides on its
next action based on the belief. As discussed earlier, we use a POMDP
for decision making. Next, we discuss parts of our system that the
POMDP requires for planning. We will discuss the actions available to
the robot, how to sample a new state, how to sample an observation,
and how to compute the observation probability.

\subsection{Actions}
In our problem setting, the robot may grasp an object and move it. We
employ top-down grasping. For selecting the finger distance and
rotation of the robot hand, we use a simple approach based on
computing a vector at a narrow part of the unknown object with
principal component analysis (PCA).
The approach 1) projects the point cloud $PC_1$ of the target object
hypothesis onto a plane, which is parallel to the wrist of the
down-pointing robot hand, 2) makes the point density of the projected
point cloud uniform to get the point cloud $PC_2$, and 3) projects the
centroid of $PC_2$ towards $PC_1$ along the wrist-plane normal, by the
distance between the centroids of $PC_1$ and $PC_2$, to get the grasp
centroid. The approach computes the PCA decomposition of $PC_2$. In
PCA, the first eigenvector aligns to the largest variance in the point
cloud data and the second eigenvector aligns along the second largest
variance in the data. For example, for a long object, the first
eigenvector could be aligned along the length of the object, and the
second eigenvector along the width of the object. The approach uses
the second eigenvector in order to get a narrow grasp. In more detail,
the approach projects two points in opposite directions from the grasp
centroid, along the computed second eigenvector, far enough. Finally,
the approach selects the two points from $PC_1$ which are closest to
the projected points, as the two grasp contact points. In addition,
the approach checks whether some part of another object hypothesis
blocks the direct path up from the two grasp contact points, and if
so, sets the probability of a successful grasp to zero.

\subsubsection{Restricted action set}
In order to restrict the computational load, we bound the number of
possible grasps, and thus actions, by a predefined maximum
number. Instead of restricting the number of possible object
hypotheses, we select a subset of all hypotheses to use for
grasping. Optimally, we would like to choose a set of grasps, which
yields the best policy among all possible grasp sets. However, because
we do not know the best policy, we settle for computing an action set
that maximizes the expected grasp success probability. The grasp
success probability $P_{\textrm{grasp prob}}(\textrm{SUCCESS} | h_i,
A)$ defines the probability of successfully grasping and moving an
object hypothesis $h_i$ when the robot chooses the best action from
the action set $A$. The expected grasp success probability is
\begin{equation}
  A = \arg\max_A \sum_{h_i}
  P_{\textrm{grasp prob}}(\textrm{SUCCESS} | h_i, A) P(h_i) \;,
\end{equation}
where the number of actions is $|A|$. $A$ can be found by an integer
linear program. Unfortunately, integer linear programming is in the
worst case NP-hard. As an approximation, we use a greedy approach
which incrementally selects object hypothesis which increase the
expected grasp success probability the most. In the experiments, the
expected grasp success probability using a restricted action set
remained usually close to the probability with the complete set of
possible actions.

\subsubsection{Grasp success probability} 
A grasp is parameterized by the distance of the finger tips, rotation
of the hand, and the location of the robot wrist. The grasp success
probability is the product of the \emph{grasp quality} and an
occlusion specific grasp probability. When computing grasp
probabilities we take previously executed grasps into account: a
failed grasp cannot succeed again, unless the occlusion of the object
changes for which the grasp is optimized (when the occlusion changes
the grasp usually changes also). Grasp quality is intended to capture
the quality of a grasp which is optimized for another object
hypothesis. Grasp quality is equal to $1$ when using a grasp which was
computed for the same object hypothesis that the robot tries to
grasp. The grasp quality decreases when the grasp centroid moves away
from the optimized grasp centroid, and becomes zero when it is outside
the object. We compute the grasp quality for grasping object
hypothesis $X$ with a grasp optimized for object hypothesis $Y$ as
follows
\begin{enumerate}
\item Inside $X$, find starting point $y_1$ and end point $y_2$ between
  the grasp points of $Y$
\item If there is no $y_1$ or $y_2$, then the grasp quality
  is zero because the line of grasp is outside object $X$
\item Compute centroid $c_Y$ of $y_1$ and $y_2$
\item Project $c_Y$ into the robot arm wrist plane along the plane normal
\item Project grasp centroid of $X$ into the wrist plane along the plane
  normal to get $c_X$
\item Denote with $\hat{X}$ the projection of $X$ onto the wrist
  plane. Project a point starting from $c_X$ through $c_Y$ so that it
  is outside $X$ and find the closest point $x_1$ in $\hat{X}$
\item Finally, compute the grasp quality as (distance from $c_Y$ to
  $x_1$) / (distance from $c_X$ to $x_1$), that is, the grasp quality
  decreases when the effective grasp centroid $c_Y$ moves closer to
  the surface, away from the grasp centroid $c_X$ optimized for $X$.
\end{enumerate}

\subsection{Temporal model of the world}
\label{sec:world_model}
In order to use the POMDP method in \cite{pajarinen2017robotic} for planning,
we need to model the evolution of the state of the world over time. We
need state transition and observation probabilities. Because
probability distributions use a state particle representation, we
need, in particular, a way to sample states and observations, and a
way to estimate the likelihood of a state particle given an
observation. Next, we discuss how to accomplish these tasks.

\emph{State sampling.} As discussed earlier, a world state consists of
an object composition $h = (h_1,\dots,h_N)$, and contains for each
$h_i$ a semantic object location, attributes, and history. To sample a
new state for a grasp action $a$, select the object hypothesis $h_i$
that has the highest grasp probability for $a$. Sample grasp success
of $a$ on $h_i$ according to the grasp success probability. If the
grasp fails, add the grasp to the grasp failure history of $h_i$. If
moving an object succeeds, the semantic location of the object is
changed to the destination location.

\emph{Observation sampling.} After executing the grasp action, the
robot observes which object was moved, and in the case of a successful
move, the robot makes an observation about the attributes (color in
the experiments) of a limited number of objects behind the moved
object. Assuming independence between attribute observations, the
observation probability is $\prod_i P(o_i|h_i)$, where $o_i$ is the
observation of $h_i$. As in \cite{pajarinen2017robotic}, $P(o_i|h_i)$ is
computed from the occlusion of $h_i$ and the attribute instances of
$h_i$.

\emph{Observation probability.} The probability of making an
observation $o$ in state $s$ is zero if the moved object hypothesis
differs from the observed one, or if the move fails and the attribute
observations do not match with previous attribute
observations. Otherwise, the probability is defined by $\prod_i
P(o_i|h_i)$ discussed above.

\section{EXPERIMENTS}
\label{sec:experiments}

The experiments are designed to test whether taking object composition
uncertainty into account improves performance in robotic
manipulation. Our hypothesis is that modeling the uncertainty
explicitly is beneficial. Furthermore, we hypothesize that planning
actions to accomplish a task under object composition uncertainty
improves performance. The two main questions we want to answer are:
\begin{enumerate}
\item Does taking segmentation uncertainty in decision making into
  account increase performance?
\item Does multi-step planning based on a distribution over object
  compositions increase performance compared to greedily choosing
  actions?
\end{enumerate}
In order to provide justified answers to these questions we compare
the baseline and proposed methods in two different tasks with two
different robot arms to show the generality of the methods. In the
first task, the robot needs to remove objects from a table using a
Kinova Jaco robotic arm. In the second task the robot needs to search
for red objects from a pile of objects in both simulation and using a
Franka Panda robot arm. In the first task we focus purely on comparing
decision making with and without taking segmentation uncertainty into
account. In the second task, multi-step planning is needed to utilize
both information gathering and task directed actions under object
composition uncertainty. Therefore, we evaluate also our POMDP based
planning method in the second task.

\subsection{Segmentation, estimating grasp probabilities}
\label{sec:experiment_methods}

In the experiments, segmentation and segment-pair probability
computation is performed using the approach presented in
\cite{richtsfeld14}. In short, the approach assigns probabilities to
segment pairs using support vector machines (SVMs) trained with RGB-D
data of household items which are not in all ways similar to the toys
we use in the experiments.

At each time step, the RGB-D sensor captures an RGB-D image of the
scene, and the system segments the RGB-D image into patches and
computes a prior probability for each patch pair to belong to the same
object using the approach in \cite{richtsfeld14}. In more detail,
\cite{richtsfeld14} groups neighbouring pixels into clusters and fits
planes and B-splines onto the patches to get parametric models (see
Fig.~\ref{fig:approach_overview} for examples of segmented
patches). \cite{richtsfeld14} computes for each patch pair a set of
features based on the texture, distance of the patches from each
other, and other properties. Finally, \cite{richtsfeld14} inputs the
computed features into a support vector machine (SVM), trained with a
labeled set of household items which differ from the items in our
experiments, and scales the output into a probability indicating
whether the patches belong to the same object. We use the approach of
\cite{richtsfeld14} to compute prior probabilities $P(c_{i,j})$ for
all segment/patch pairs, where $P(c_{i,j} = 1)$ defines the prior
probability for $i$ and $j$ to be part of the same object. Note that,
in place of the segmentation approach we currently use
\cite{richtsfeld14}, our approach can use also other approaches to
over-segment and estimate patch pair probabilities.

We use $P(c_{i,j})$ in the MCMC procedure in
Algorithm~\ref{alg:sample_compositions} and compute a probability
distribution over object compositions (see
Fig.~\ref{fig:approach_overview} for examples). The number of MCMC
samples should be chosen according to the computational budget. In the
experiments, we used $\mathbf{|H|} = 2000$ samples. Because the
minimum lower bound estimate ESS underestimates the real ESS it should
be lower than the number of samples: we used a target ESS of
$N_{\textsc{ESS}} = 200$, a tenth of the number of samples. The number
of CFTP starting states influences the independence of the first
sample w.r.t.\ the starting state. We used $N_{\textsc{start}} = 100$
CFTP starting states in the experiments. The hard limit for the number
of samples generated was $T_{\textsc{max}} = 131072$.

\textbf{Grasp probabilities.} For grasp probability we used parameters
estimated for coffee cups in \cite{pajarinen2017robotic} and we set
the first color observation parameter (see \cite{pajarinen2017robotic}
for details) to $-0.5$ and the second to $-0.02$ for both red and
non-red observations. The models were not optimized for the particular
objects because in the real-world the robot would need to be able to
generalize to new objects.

\subsection{Experiments: Clearing a Table}
\label{sec:clearing_table}

\begin{figure}[thpb]
    \centering 
    \setlength{\tabcolsep}{2pt}
    \begin{tabular}{cc}
    \includegraphics[height=2.9cm]{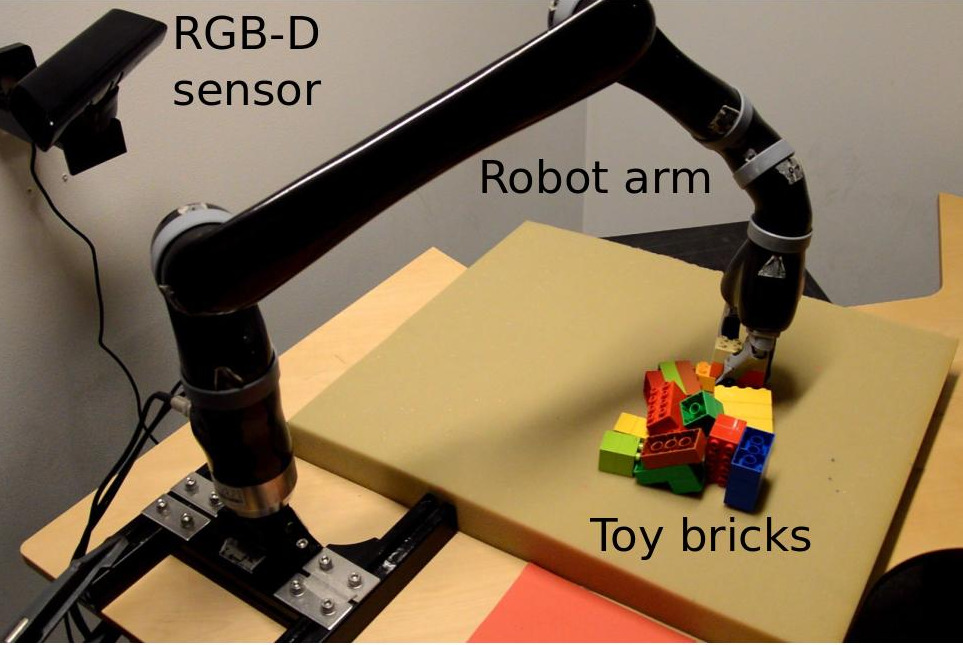}&
    \includegraphics[height=2.9cm]{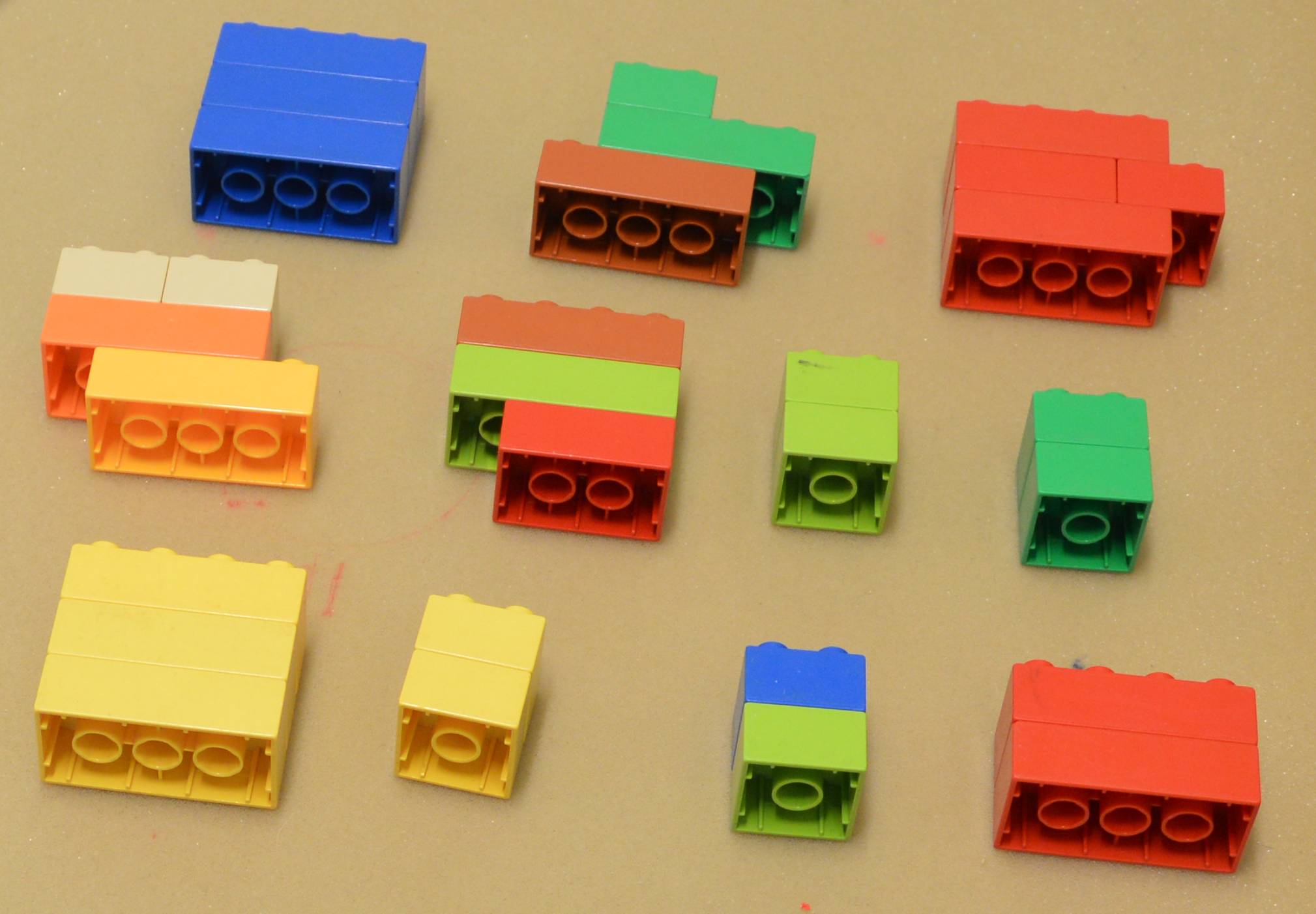}
    \end{tabular}
    \caption{In table clearing, we use an RGB-D sensor for visual input
        and a 6-DOF Kinova Jaco arm for grasping randomly placed toy
        bricks. \textbf{(Left)} Overall experimental setup.
        \textbf{(Right)} Toy bricks used in the experiment.}
    \label{fig:setup_jaco}
\end{figure}

Fig.~\ref{fig:setup_jaco} shows the experimental setup for table
clearing. In the setup, a Kinect RGB-D sensor captures images of the
scene and a 6-DOF robotic Kinova Jaco arm tries to move as many toy
bricks away from the table as possible. Since we do not assume any
prior information in advance, including geometric or colour
information, and because the bricks are in a pile, segmenting the
bricks correctly is difficult. For clearly separated known objects one
could possibly use standard segmentation methods.

\begin{figure*}[tb]
  \centering
  \vspace{0.4em}%
  \subfloat[Random scenes in ``Best segmentation'' evaluations, ordered according to experimental success from best to worst]{\label{fig:random_lego_scenes_A}%
    \begin{tabular}{lllll}
      \includegraphics[width=3.1cm]{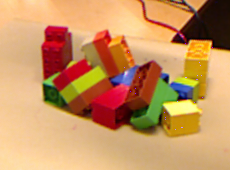} &
      \includegraphics[width=3.1cm]{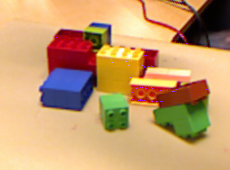} & 
      \includegraphics[width=3.1cm]{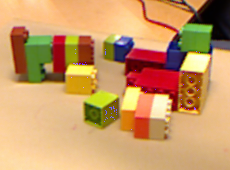} &
      \includegraphics[width=3.1cm]{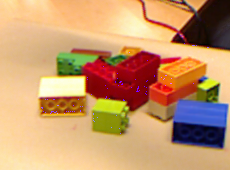} &
      \includegraphics[width=3.1cm]{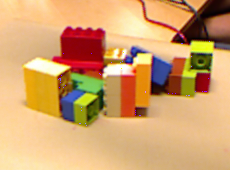} \\
      \includegraphics[width=3.1cm]{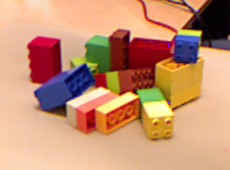} &
      \includegraphics[width=3.1cm]{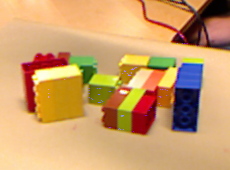} &
      \includegraphics[width=3.1cm]{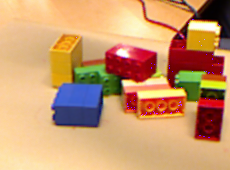} &
      \includegraphics[width=3.1cm]{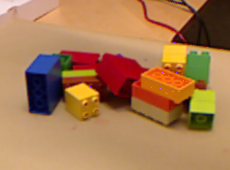} &
      \includegraphics[width=3.1cm]{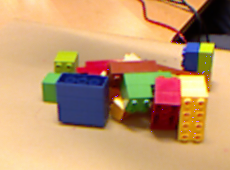}
    \end{tabular}}\\
  \subfloat[Random scenes in ``Maximum utility'' evaluations, ordered according to experimental success from best to worst]{\label{fig:random_lego_scenes_B}%
    \begin{tabular}{lllll}
      \includegraphics[width=3.1cm]{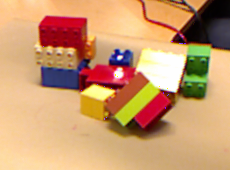} &
      \includegraphics[width=3.1cm]{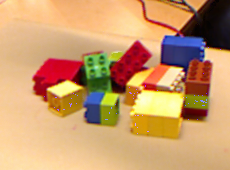} & 
      \includegraphics[width=3.1cm]{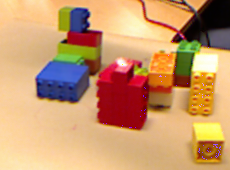} &
      \includegraphics[width=3.1cm]{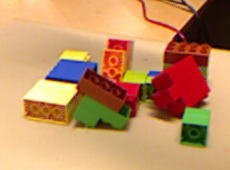} &
      \includegraphics[width=3.1cm]{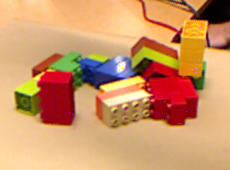} \\
      \includegraphics[width=3.1cm]{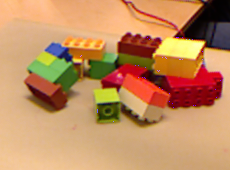} &
      \includegraphics[width=3.1cm]{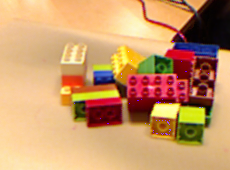} &
      \includegraphics[width=3.1cm]{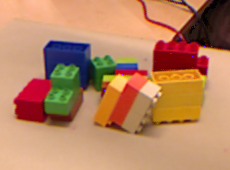} &
      \includegraphics[width=3.1cm]{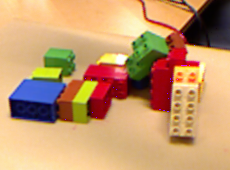} & 
      \includegraphics[width=3.1cm]{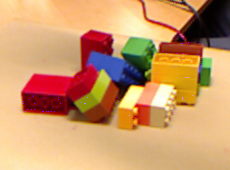}
    \end{tabular}}
  \caption{Cropped kinect RGB images of the 20 randomly generated
    scenes in table clearing.}
  \label{fig:random_lego_scenes}
\end{figure*}

In the experiments, we shook a box containing toy bricks shown in
Fig.~\ref{fig:setup_jaco} and emptied the bricks into a specific area
on a table. Fig.~\ref{fig:random_lego_scenes} shows the 10 random
scenes for each method ordered so that the first scene produced
highest utility and the last scene the lowest utility. The goal was to
move a toy brick in each time step away from the table. In the
application, action $a$ specifies the object to grasp and move
away. The robot is rewarded $1$ for a successfully grasped and moved
object and $0$ otherwise.

Instead of POMDP planning we optimize in this experiment only expected
immediate reward to evaluate whether utilizing a probability
distribution over object compositions increases performance compared
to utilizing only the best object composition found. We compare two
methods. The first one, called ``Best segmentation'' finds first the
most likely object composition $\mathbf{h^*}$, and then finds the
action $a$ that maximises the utility or immediate reward function
$R(\mathbf{h^*}, a)$. In table clearing, ``Best segmentation'' tries
to grasp the object which has the highest grasp success probability in
the most likely object composition. The second method, called
``Maximum utility'', corresponds to maximizing the expected immediate
reward where the expectation is taken over possible object
compositions. In table clearing, ``Maximum utility'' tries to grasp
the object which has the highest grasp success probability weighted by
the probability of the object to exist in an object composition.

\paragraph{Table Clearing: Results \& Discussion}

\begin{figure}[tb]
  \centering
  \vspace{0.4em}%
  \includegraphics[width=6cm]{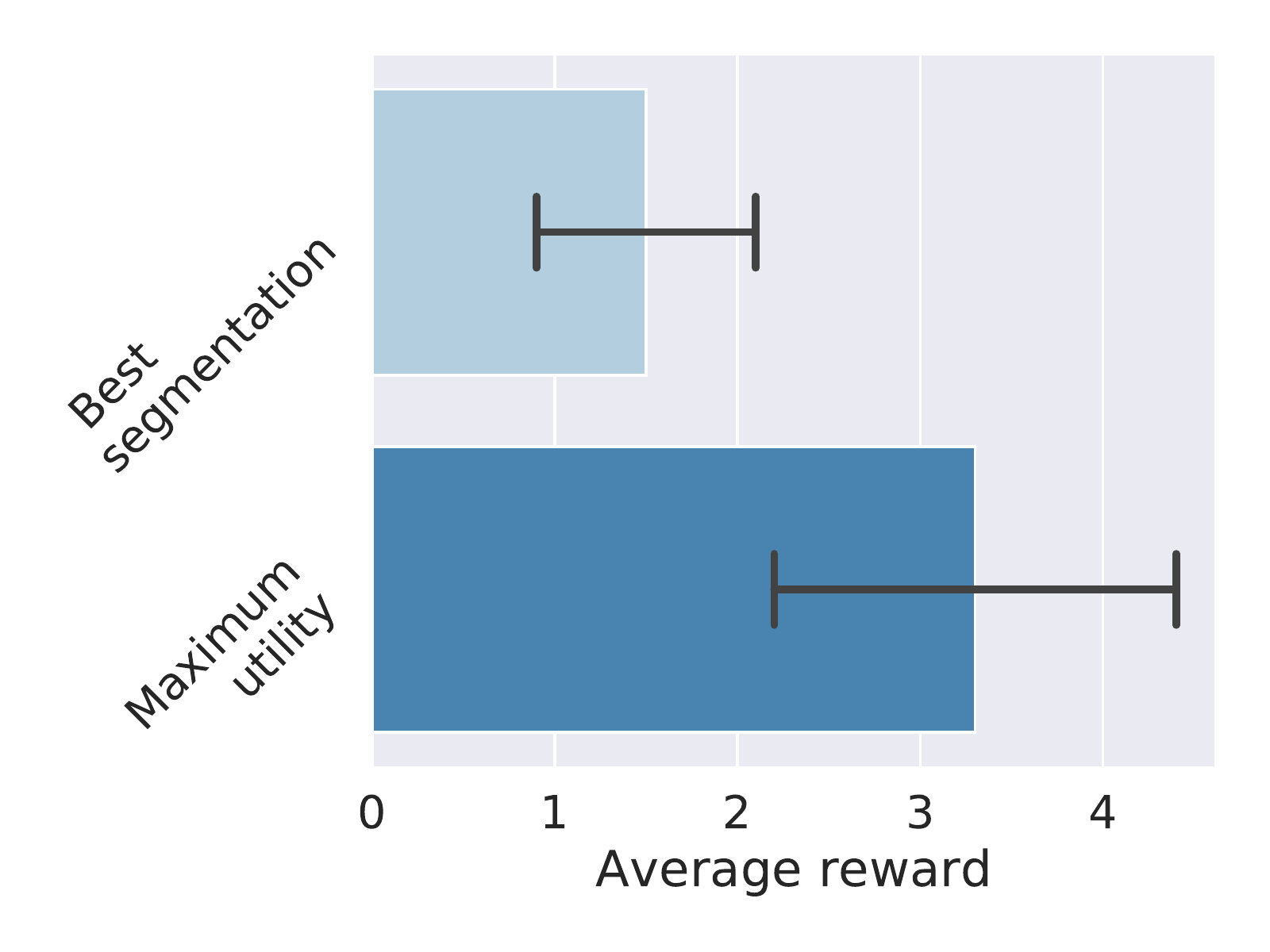}
  \caption{Results in table clearing. The robot grasps and moves toy bricks away from the table.
    The bar plot shows the average number of successful moves (average reward) 
    and the 95\% confidence interval
    in 10 experimental runs for each method.
    The ``Maximum utility'' method performed
    significantly better than ``Best segmentation'' 
    ($p=0.014$ in the Mann-Whitney $U$ one-sided test \cite{mann47}).}
  \label{fig:random_lego_results}
\end{figure}

\begin{figure}[tb]
  \centering
  \vspace{0.6em}%
  \subfloat[Segmented patches]{\label{fig:results_segmented}%
    \fbox{\includegraphics[width=3.9cm]{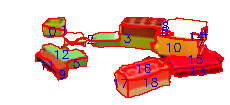}}}
  \subfloat[Most likely composition]{\label{fig:results_composition}%
    \fbox{\includegraphics[width=3.9cm]{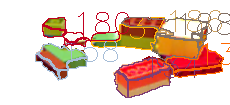}}}
  \caption{``Best object'' is able to move an object when ``Best
    composition'' fails. Time step 6 in the sixth scene in
    Fig.~\ref{fig:random_lego_scenes_B}:
    \protect\subref{fig:results_segmented} Segmented patches,
    \protect\subref{fig:results_composition} the most likely object
    composition (probability $0.271$). ``Best object'' grasps
    successfully an object hypothesis consisting of patches $0$ and
    $1$. However, ``Best composition'' finishes because segments $0$,
    $1$, and $2$ form in \protect\subref{fig:results_composition}
    object hypothesis $180$, which can not be grasped.}
  \label{fig:result_analysis}
\end{figure}

Fig.~\ref{fig:random_lego_results} shows the
number of successful moves (a maximum of six moves per scene) in 10
experimental runs for each method. One complete object movement
operation, including image processing, segmenting, generating the
particle based probability distribution, and moving an object, took on
the average $79.9s$ of which our MCMC approach took $8.8s$
($11$\%). The time needed for MCMC depends on the number of particles
and CFTP starting states and can be adjusted. The ``Maximum utility''
method performed significantly better than ``Best segmentation'' (the
$p$-value was $0.014$ in the Mann-Whitney $U$ one-sided test \cite{mann47}). To
qualitatively compare the two methods we recorded decisions by both,
although only one method operated the robot arm in each scene, that
is, we ran one method and at the same time output the decisions which
the other method would have made for the same object compositions. In
the scenes in Fig.~\ref{fig:random_lego_scenes_A}, even though
graspable objects were still available, ``Maximum utility'' would have
finished execution early $3$ times and ``Best segmentation'' finished
early $10$ times, that is, in every scene, and in the scenes in
Fig.~\ref{fig:random_lego_scenes_B} ``Maximum utility'' finished execution
early $3$ times and ``Best segmentation'' would have finished early
$23$ times. The most likely composition was often missing graspable
objects that were part of other object
compositions. Fig.~\ref{fig:result_analysis} shows an example of one
such situation.  Fig.~\ref{fig:results_segmented} shows
under-segmentation happening sometimes. In general, it is better to
over-segment too heavily than under-segment but this applies to all
over-segmentation approaches including the over-segmentation approach
utilised by the two comparison methods. Grasps were sometimes
successful even when the segmentation of the grasped object did not
correspond to a real object. For example, the robot sometimes grasped
the segmented top of an object and moved the complete object
successfully. The robot can achieve higher performance because our
utility function did not include unnecessary constraints. For
applications, such as moving fragile objects, the utility function can
penalize grasping an incorrectly segmented object if this could lead
to dropping the object.

\subsection{Experiments: Object Search}
\label{sec:object_search}

\begin{figure}[thpb]
  \centering 
  \setlength{\tabcolsep}{2pt}
  \begin{tabular}{ccc}
  \includegraphics[height=4.0cm]{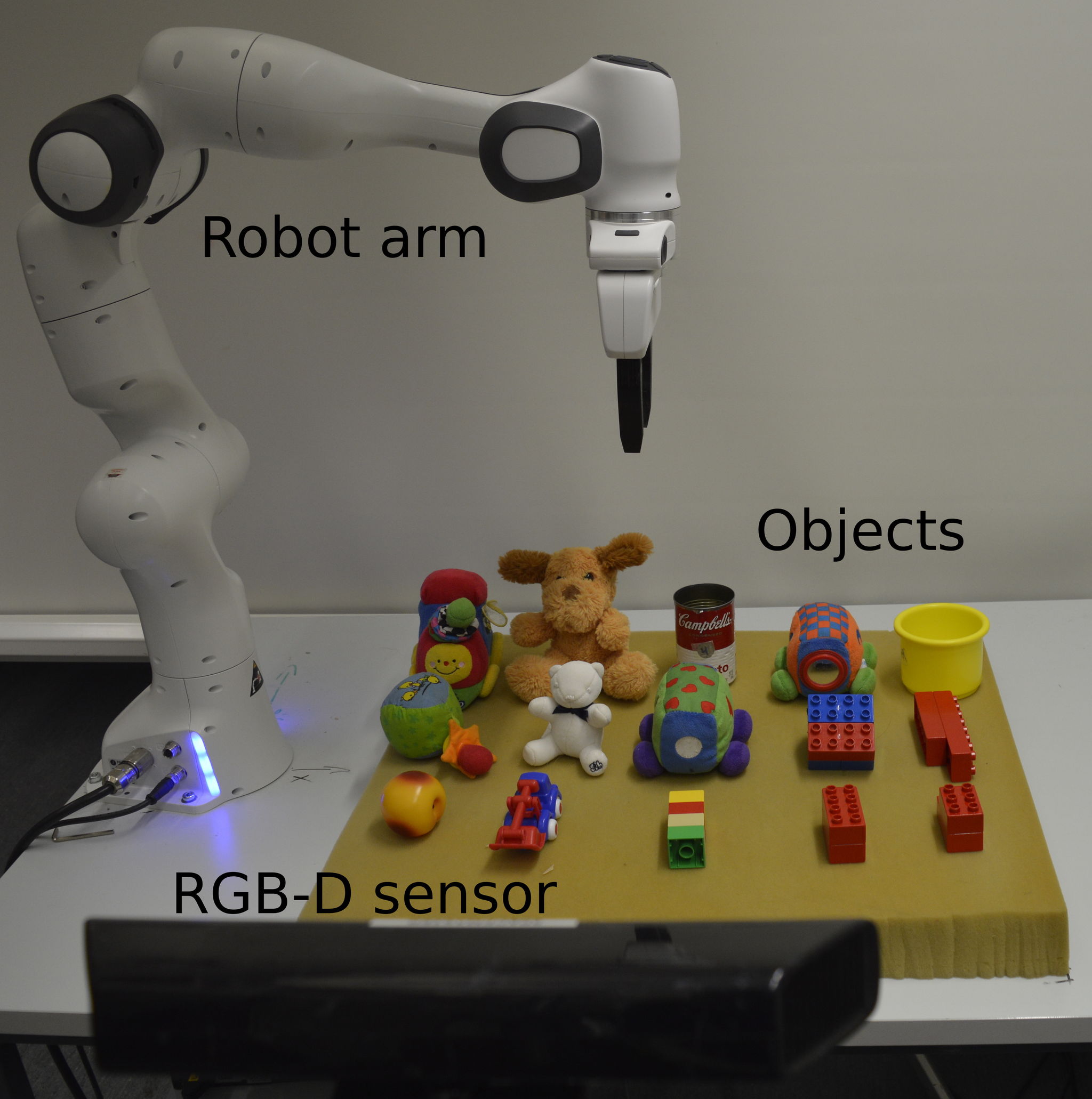}&
  \includegraphics[height=4.0cm]{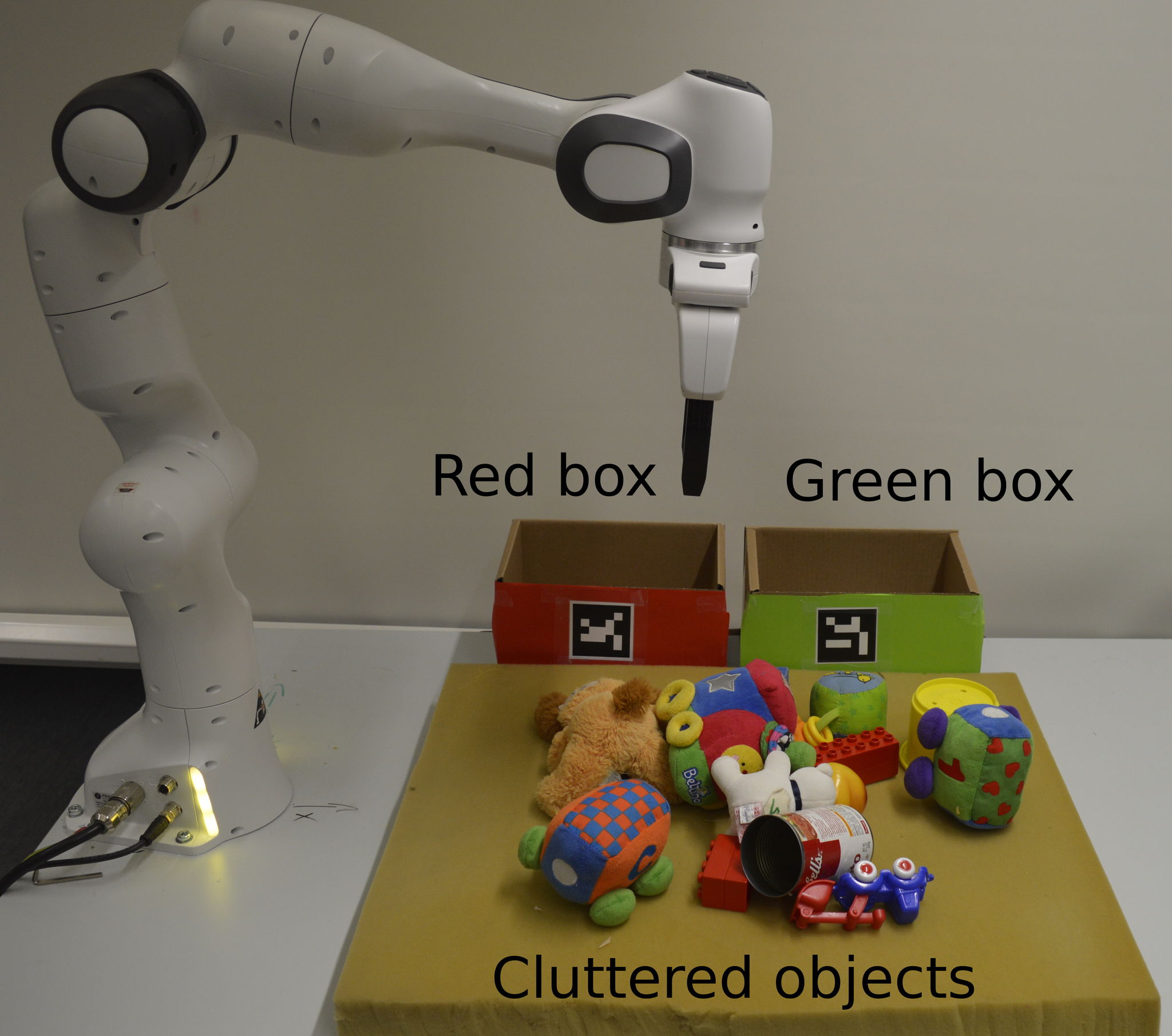}
  \end{tabular}
  \caption{Experimental setup for object search. Based on RGB-D data from a Microsoft
    Kinect, a 7-DOF Franka Panda robotic arm tries to move red objects
    into a red box. \textbf{(Left)} Robotic setup with all objects arranged on the table.
    \textbf{(Right)} In actual experiments, objects are dumped into the workspace resulting
    in a cluttered scene. The robot tries to move fully red objects into the red box and
    may put objects into the green box to reduce clutter.}
  \label{fig:setup_franka}
\end{figure}

In the object search experiments, a Kinect RGB-D sensor observes
objects on a table and a \panda robot arm with an attached custom
parallel gripper manipulates the objects.
Figure~\ref{fig:setup_franka} shows the objects used in the
experiment. We chose these objects as they differ in size, rigidness,
color, and texture. In order to get diverse and unbiased scenes for
each trial, we placed the objects inside a box which was shaken and
emptied into the workspace of the robot. If an object ended up outside
the workspace the process was repeated.

The task the robot had to solve was to find and move an unknown number
of fully red objects into the red box shown in
Figure~\ref{fig:setup_franka}. For every red object the robot places in the
red box we increase the score (utility) by 1 and for every non-red
object placed in the red box we decrease the score by 1. To remove
occlusions that hinder color detection and grasping, the robot may
also move objects into a green box without any direct effect on the
total score, \ie we neither add nor subtract points for such an
action.

Object search~\cite{xiao2019online,danielczuk2020x} is a well known
task in robotics but existing approaches do not explicitly take
composition uncertainty in planning actions into account. Of recent
work \cite{danielczuk2020x} performs object search for a pre-specified
target by segmenting an image in each time step and combining the
segmented image with an ``occupancy distribution'' that describes
where the object could be hidden. We consider a more general case
without a specific target object and without knowing the sizes of the
objects we are searching for. \cite{xiao2019online} uses a POMDP model
for object search but assumes perfect object composition segmentation
with some uncertainty in object locations and utilizes a priori
knowledge of number and models of objects for finding hidden objects.
We evaluated four different methods:
\begin{itemize}
\item Best grasp for objects observed red in the most likely segmentation (``Best segmentation'')
\item Grasp for the highest expected immediate reward (``Maximum utility'')
\item POMDP multi-step planning without hallucinating hidden objects (``POMDP'')
\item POMDP multi-step planning with hallucinating hidden objects (``POMDP with hallucination'')
\end{itemize}
The first two methods are similar to the ones used in table clearing
in Section~\ref{sec:clearing_table}. The ``POMDP'' method uses a
dynamics and observation model and using the model and the probability
distribution based sufficient statistic plans multi-step actions as
described in Section~\ref{sec:manipulation_planning}. The ``POMDP with hallucination''
method extends the third method with a probability model for hidden
objects as defined in Section~\ref{sec:hallucination} while the probability for
hidden objects to be red was estimated as the average fraction of
visible red objects linearly scaled into $[0.2, 0.8]$ to always allow
for both non-red and red hidden objects.

\subsubsection{Object Search with Simulated Dynamics.}
\label{sec:experiments_franka_sim}

\begin{figure}[tb]
    \centering
    \vspace{0.4em}%
    \includegraphics[width=7cm]{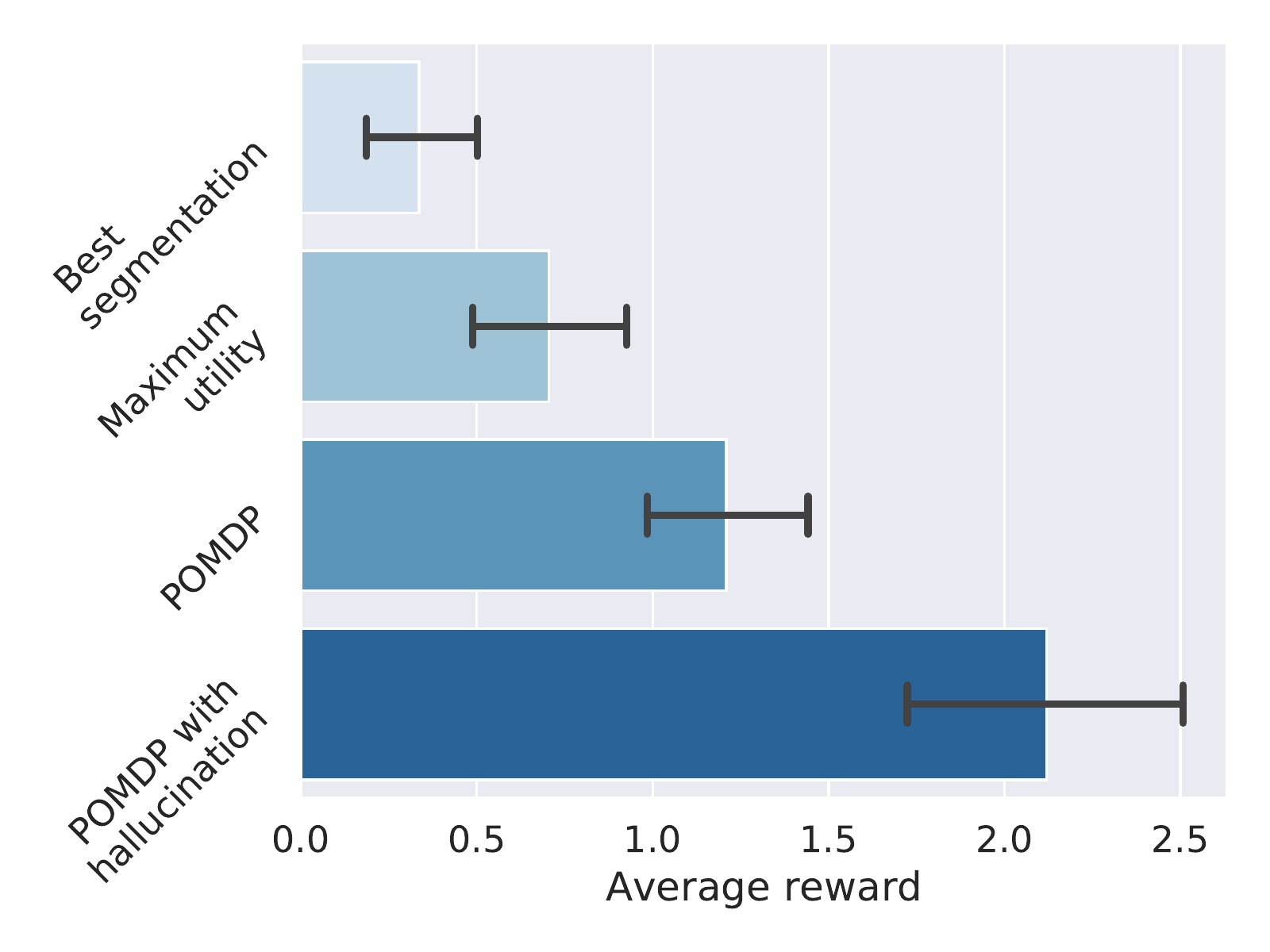}
    \caption{Simulated object search.
        The bar plot shows the average reward
        and the 95\% confidence interval for each method. 
        The differences between the method performances is statistically significant.
        Please, see the main text for further discussion.}
    \label{fig:franka_sim}
\end{figure}

In object search simulation experiments, we used the initial RGB-D
scenes captured in the robot experiments discussed in
Section~\ref{sec:experiments_franka_real} but simulated dynamics and
observation probabilities. Fig.~\ref{fig:franka_sim} shows the
results. For statistical analysis we ran a Kruskal-Wallis test and
then Posthoc Conover tests revealing that all methods' average reward
differed statistically significantly: "POMDP with hallucination" from
"POMDP" ($p=0.002$), "POMDP" from "Maximum utility" ($p=0.003$), and
"Maximum utility" from "Best segmentation" ($p=0.021$).

\subsubsection{Object Search on Real Hardware.}
\label{sec:experiments_franka_real}

\begin{figure}[thpb]
	\centering 
	\setlength{\tabcolsep}{2pt}
	\begin{tabular}{cccc}
		\includegraphics[height=2.9cm]{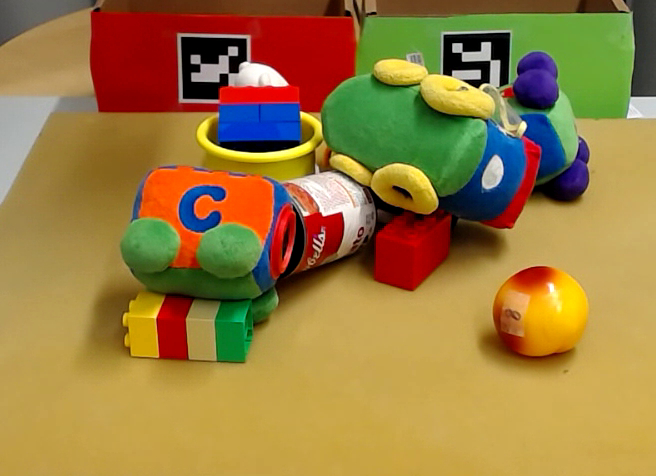}&
		\includegraphics[height=2.9cm]{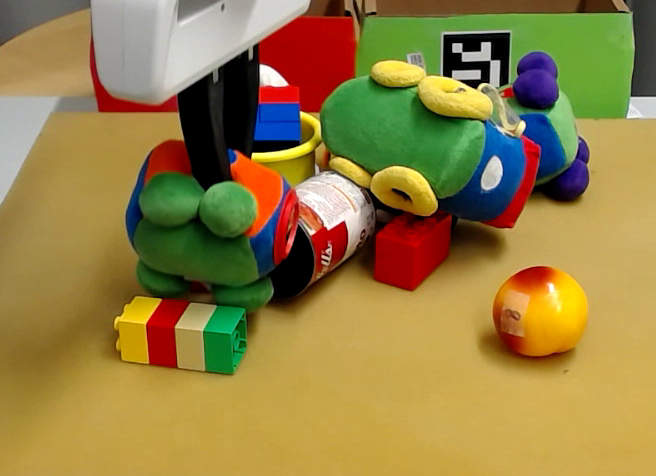}&
		\\
		\includegraphics[height=2.9cm]{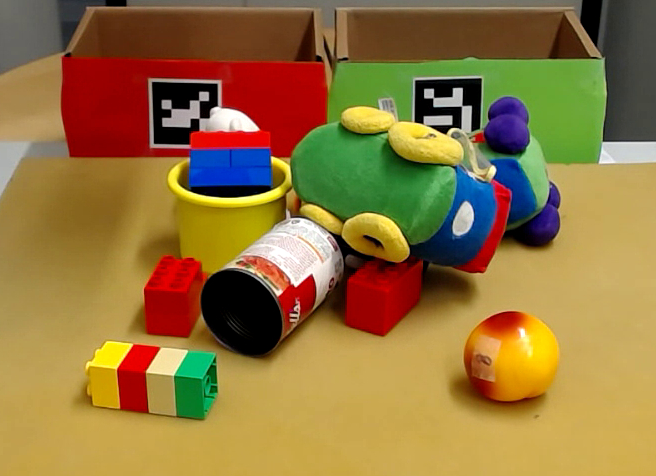}&
		\includegraphics[height=2.9cm]{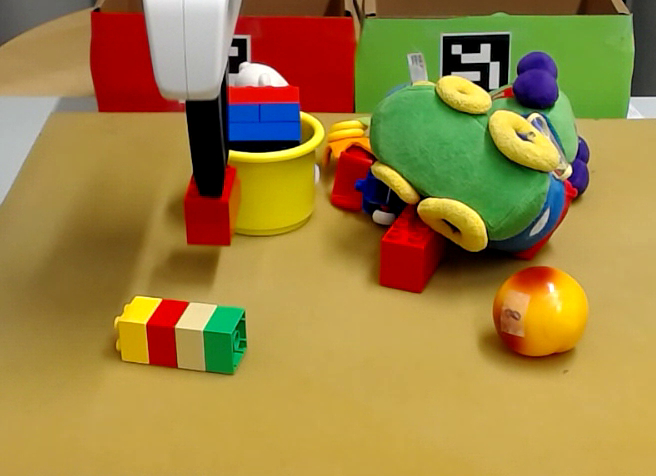}
	\end{tabular}
	\caption{The "POMDP with hallucination" method removes objects
          occluding a large space to reveal a fully red
          object. \textbf{(Upper Left)} Initial scene. \textbf{(Upper
            Right)} Picking up non-red object occluding a large
          space. \textbf{(Bottom Left)} A fully red hidden object is
          revealed. \textbf{(Bottom Right)} The robot now picks up the
          red object and moves it into the red box.}
	\label{fig:example_scene}
\end{figure}

\begin{figure}[tb]
    \centering
    \vspace{0.4em}%
    \includegraphics[width=7cm]{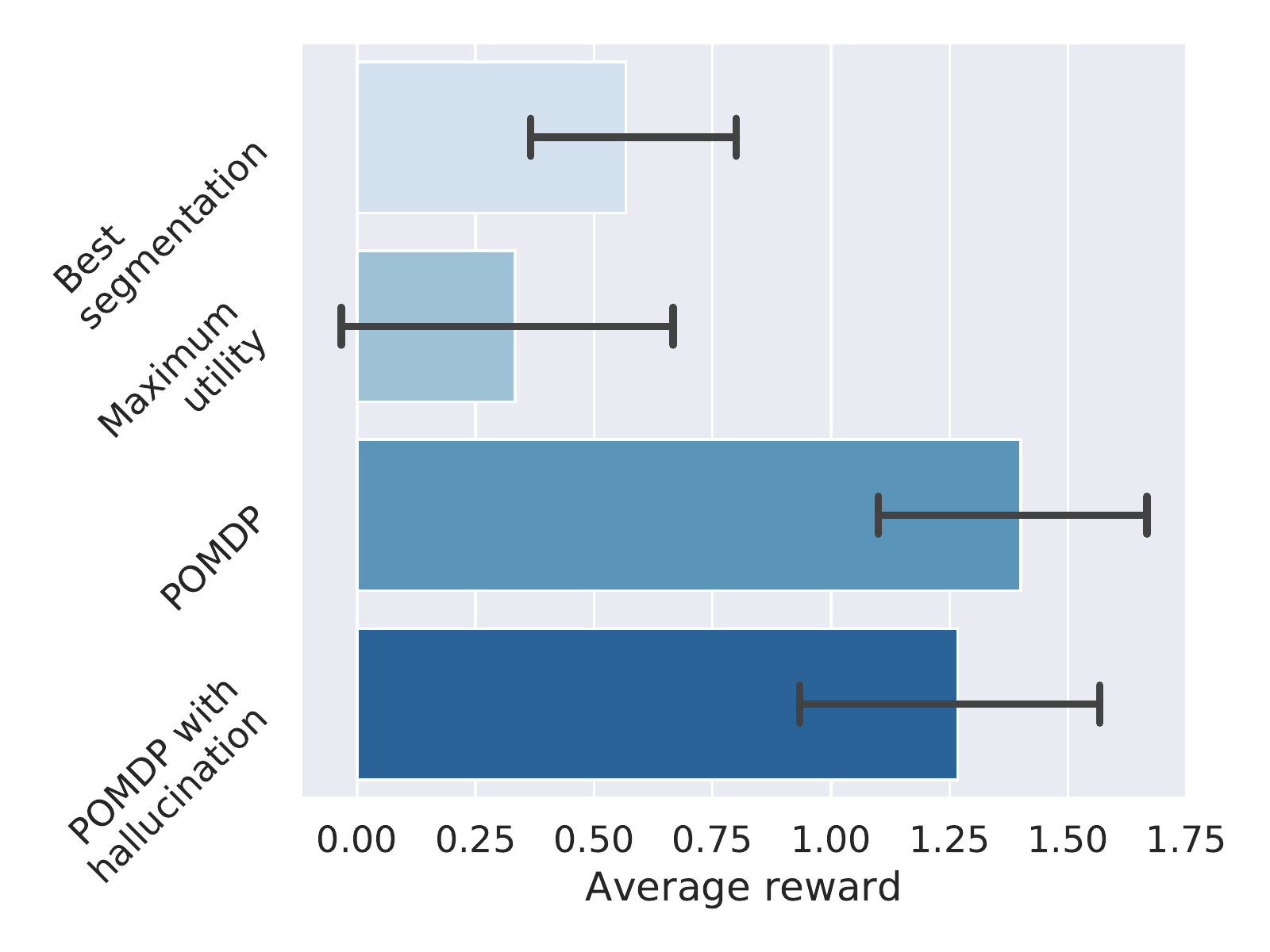}
    \caption{Object search with a Franka Panda robot arm.
        The bar plot shows the average reward
        and the 95\% confidence interval for each method. 
        The performance difference between the POMDP methods and the 
        greedy methods is statistically significant.
        Please, see the main text for further discussion.}
    \label{fig:franka_real}
\end{figure}

Fig.~\ref{fig:setup_franka} shows the experimental setup and one of
the scenes we used for object search with a Franka Panda robot. As in
previous experiments, we randomly generated scenes by adding all
objects into a box, shook the box, and emptied the content into the
robot's workspace.

\begin{figure}[thpb]
  \centering 
  \setlength{\tabcolsep}{2pt}
  \begin{tabular}{cccc}
    \includegraphics[height=1.2cm]{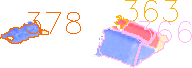}&
    \\
    \includegraphics[height=4.9cm]{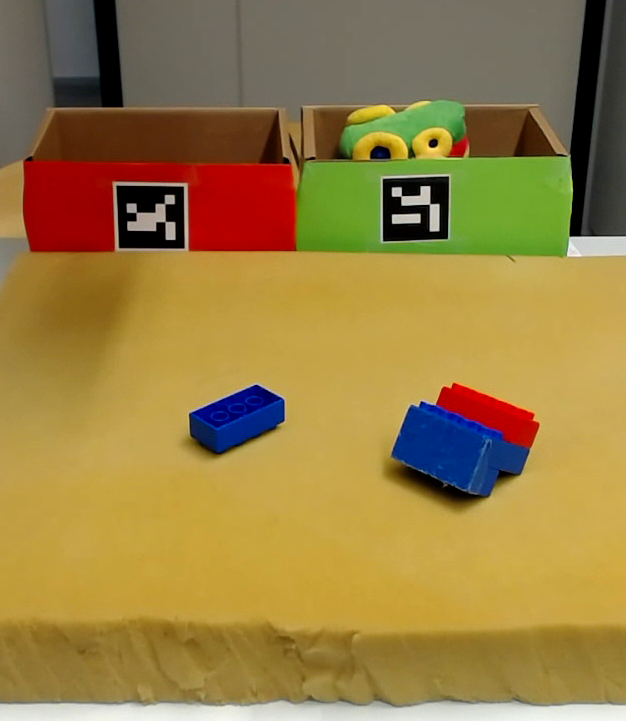}&
    \includegraphics[height=4.9cm]{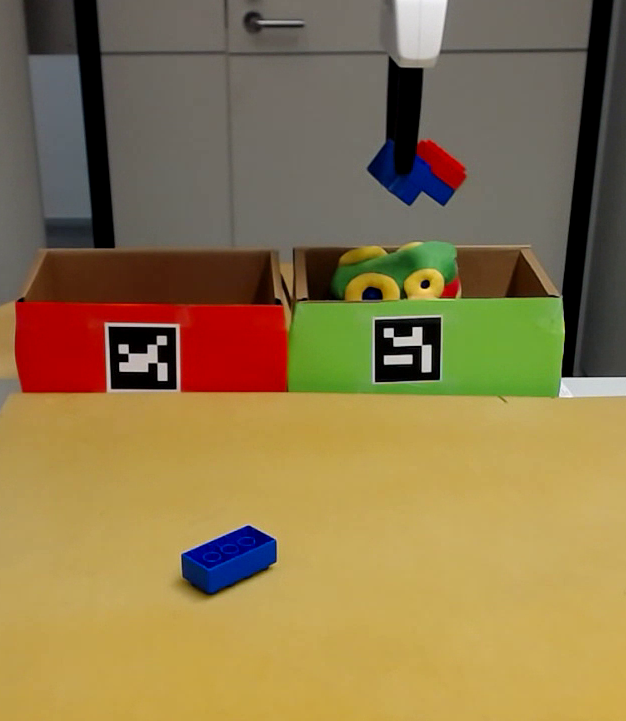}
  \end{tabular}
  \caption{\textbf{(Left)} The object composition on the top indicates
    that the red lego block is either separated or part of the larger
    blue object. A reasonable approach here is to grasp the larger
    blue part and move it to the green box. Then if the composition is
    separable the red lego will be left in the scene while if the red
    is part of the larger blue object the red-blue object will
    disappear which is also fine. \textbf{(Right)} Robot picked up the
    larger blue object. In this case the red segment was part of the
    blue object.}
  \label{fig:smart_pomdp}
\end{figure}

\begin{figure}
  \centering
  \includegraphics[width=7cm]{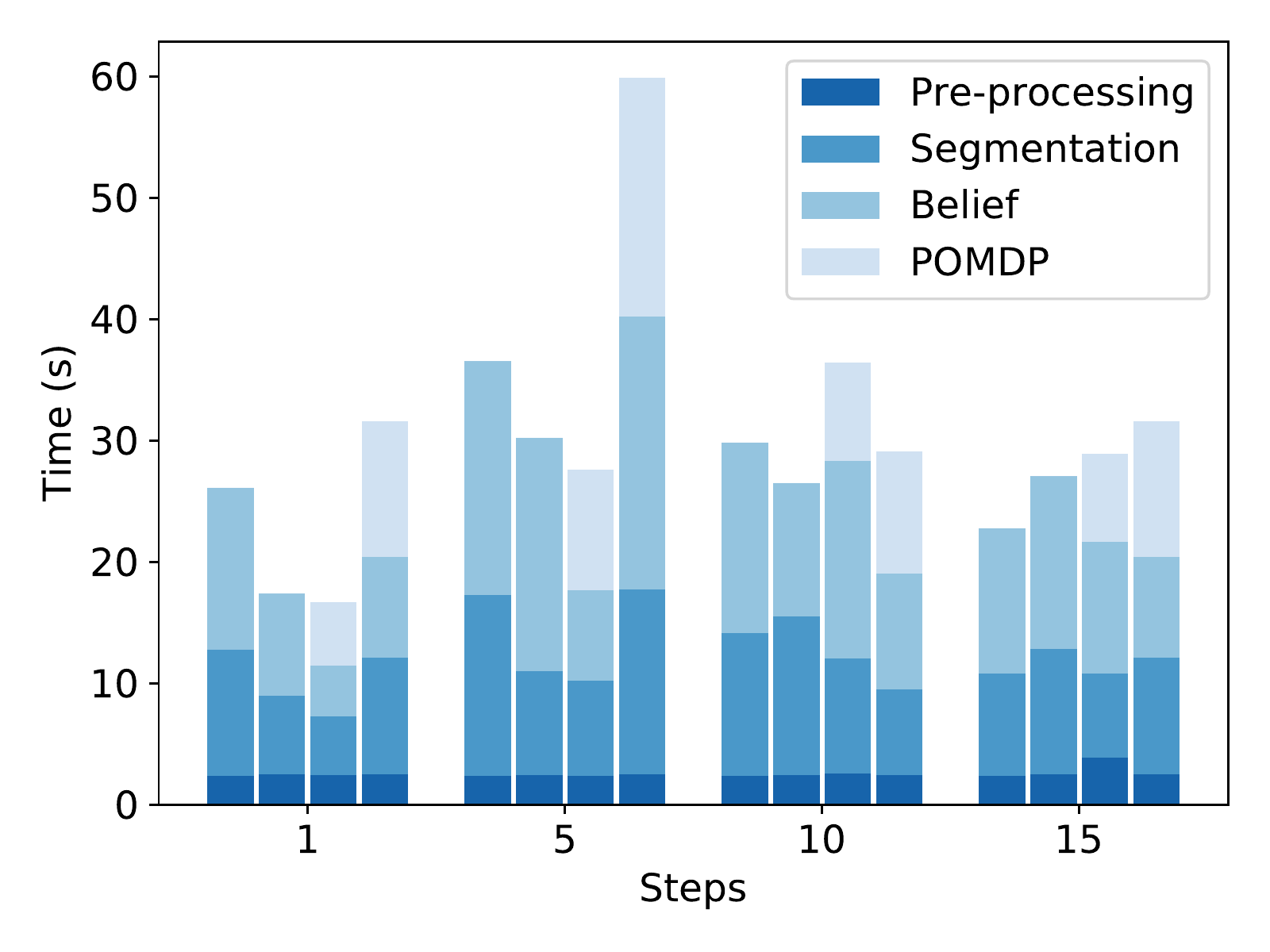}
  \caption{The average running time for each comparison method at
    different phases of the manipulation task, that is, the time step.
    The running time is split into pre-processing, segmentation,
    belief generation, and the POMDP solver components. The four
    comparison methods are, from left to right,``Best segmentation'',
    ``Maximum utility'', ``POMDP'' and ``POMDP with
    hallucination''. Segmentation, belief generation, and the POMDP
    solver take roughly the same amount of processing time. This is by
    design since the running time of belief generation and the POMDP
    solver can be controlled while increasing or decreasing
    the exactness of the computations.}
  \label{fig:running_times}
\end{figure}

Fig.~\ref{fig:franka_real} summarizes the results: POMDP methods
significantly outperformed the greedy approaches. The main reasons the
POMDP methods outperform the greedy approaches is that they utilize
information gathering actions and remove occluding non-red objects as
shown in Fig.~\ref{fig:example_scene} and plan actions over
distributions of compositions. Planning over a distribution of
compositions discourages our approach from stopping prematurely,
contrary to approaches that may stop when no red object is seen in the
most probable distribution. Moreover, planning using a distribution of
compositions enables complex reasoning such as moving a non-red object
hypothesis away to verify whether it forms together with another red
object hypothesis an object as shown in Fig.~\ref{fig:smart_pomdp}: if 
the hypotheses do not belong together
we have removed a non-red object and can next collect the red object,
while if the red and non-red hypotheses are part of the same object we
still have removed only a non-red object and clarified the situation.
Fig.~\ref{fig:running_times} shows the running times for all
comparison methods for the different computational components at
different phases of the manipulation task.

``POMDP with hallucination'' removed 21 occluding non-red objects to
reveal a fully red object and ``POMDP'' 18 occluding non-red objects.
Fig.~\ref{fig:example_scene} shows one successful example where
``POMDP with hallucination'' removes an obstacle to reveal a fully red
object which is subsequently moved into the red box. We hypothesize
that with bigger and/or more occluding non-red objects or fewer red
objects in total the performance of ``POMDP with hallucination'' may
outperform the regular ``POMDP'' method as it has an inherent
advantage of removing occluding objects in the scene.

\section{CONCLUSIONS}
\label{sec:conclusions}

Manipulating unknown objects in a cluttered environment is a hard but
important problem in robotics. Allowing robots to operate in cluttered
unknown environments is essential for robot autonomy and succeeding in
tasks such as waste segregation, agile manufacturing, service
robotics, and rescue robotics. However, a lack of object models and a
noisy partial view make manipulation in such environments
difficult. To succeed in such tasks the robot needs to consider
possible object compositions and in particular plan actions so that
the actions chosen take all possible object compositions now and in
future time steps into account. Therefore, instead of utilizing only
the most likely object composition, we plan manipulation actions in
the space of object compositions. In object search and table clearing
experiments, our approach outperforms an approach based on the most
likely object composition. Moreover, long term planning outperforms a
greedy approach when planning over a distribution over object
compositions.

In this paper we considered a setting with static objects. Future work
includes manipulation of dynamic moving objects under
occlusion. Moreover, our probabilistic world model allows for
straightforward inclusion of prior knowledge when available and thus
inclusion of different types of sensors such as tactile sensors is an
avenue for future work. We expect the main idea of planning behavior
based on a distribution of object compositions to transfer to many
application domains. For example, in autonomous driving there can be
high uncertainty due to occlusion and weather conditions: which parts
of the scene are pedestrians?, which parts are cars?, which are
buildings? Planning based on the distribution of object compositions
is a safer alternative than relying on the most likely object
composition which may be incorrect.


%%%%%%%%%%%%%%%%%%%%%%%%%%%%%%%%%%%%%%%%%%%%%%%%%%%%%%%%%%%%%%%%%%%%%%%%%%%%%%%%

%%%%%%%%%%%%%%%%%%%%%%%%%%%%%%%%%%%%%%%%%%%%%%%%%%%%%%%%%%%%%%%%%%%%%%%%%%%%%%%%



% \section*{ACKNOWLEDGMENT}

%%%%%%%%%%%%%%%%%%%%%%%%%%%%%%%%%%%%%%%%%%%%%%%%%%%%%%%%%%%%%%%%%%%%%%%%%%%%%%%%

% References are important to the reader; therefore, each citation must
% be complete and correct. If at all possible, references should be
% commonly available publications.

\bibliographystyle{IEEEtran}
\bibliography{root}

\end{document}